\documentclass{article}

\usepackage[accepted]{aistats2019}
\usepackage{xr}
\usepackage[utf8]{inputenc} 
\usepackage[T1]{fontenc}    
\usepackage[hidelinks]{hyperref}       
\usepackage{url}            
\usepackage{booktabs}       
\usepackage{amsfonts}       
\usepackage{nicefrac}       
\usepackage{microtype}      

\usepackage{natbib}
\usepackage{rotating}
\usepackage{subfig}
\usepackage{graphicx}
\usepackage{sidecap}
\usepackage{JalilMath}
\usepackage{amssymb}
\usepackage{tikz}
\usepackage{multirow}
\usetikzlibrary{fit,positioning}
\usepackage{amsfonts}
\usepackage{graphicx,colonequals}
\usepackage[inline]{enumitem}
\usepackage[toc]{appendix}
\usetikzlibrary{shapes}

\newcommand{\h}[1]{^{(#1)}}

\newcommand{\la}{\left\langle}
\newcommand{\ra}{\right\rangle}
\newcommand{\TT}{\mathcal{T}}

\newcommand{\yt}{\vc y_{\mathcal{T}}}
\newcommand{\zt}{\vc z_{\mathcal{T}}}
\newcommand{\xt}{\vc x_{\mathcal{T}}}
\newcommand{\xtj}[1]{\vc x_{\mathcal{T}^{(#1)}}}
\newcommand{\xtjl}[2]{\vc x_{\mathcal{T}_{#2}^{(#1)}}}
\newcommand{\ytj}[1]{\vc y_{\mathcal{T}^{(#1)}}}
\newcommand{\ztj}[1]{\vc z_{\mathcal{T}^{(#1)}}}

\newcommand{\ftj}[1]{\bc f_{\mathcal{T}^{(#1)}}}

\newcommand{\ztjl}[2]{\vc z_{\mathcal{T}_{#2}^{(#1)}}}

\newcommand{\ftjl}[2]{\bc f_{\mathcal{T}_{#2}^{(#1)}}}

\newcommand{\res}{^{(j)}}

\newcommand{\bc}{\mathbf}

\usepackage{natbib}
\PassOptionsToPackage{numbers}{natbib}
\setcitestyle{numbers,open={[},close={]}}
\title{Conditionally Independent \\ Multiresolution Gaussian Processes}

\begin{document}


\twocolumn[

\aistatstitle{Conditionally Independent Multiresolution Gaussian Processes}

\aistatsauthor{ Jalil~Taghia \And Thomas~B.~Sch{\"o}n}

\aistatsaddress{Department of Information Technology \\ Uppsala University, Sweden\\ \texttt{jalil.taghia@it.uu.se}\\
   \And  Department of Information Technology \\ Uppsala University, Sweden\\
  \texttt{thomas.schon@it.uu.se} } ]

\begin{abstract}
\vspace{-2ex}
The multiresolution Gaussian process (GP) has gained increasing attention as a viable approach towards improving the quality of approximations in GPs that scale well to large-scale data. Most of the current constructions assume full independence across resolutions. This assumption simplifies the inference, but it underestimates the uncertainties in transitioning from one resolution to another. This in turn results in models which are prone to overfitting in the sense of excessive sensitivity to the chosen resolution, and predictions which are non-smooth at the boundaries. 
Our contribution is a new construction which instead assumes \textit{conditional independence }among GPs across resolutions. 
We show that relaxing the full independence assumption enables robustness against overfitting, and that it delivers predictions that are smooth at the boundaries. Our new model is compared against current state of the art on 2 synthetic and 9 real-world datasets. In most cases, our new conditionally independent construction performed favorably when compared against models based on the full independence assumption. In particular, it exhibits little to no signs of overfitting.
\end{abstract}
\vspace{-2ex}
\section{INTRODUCTION}
\label{sec:intro}
\vspace{-2ex}
There is a rich literature on methods designed to avoid the computational bottleneck incurred by the vanilla Gaussian process (GP), including sub-sampling \citep{Rasmussen2006}, low rank approximations \citep{Cressie2008}, covariance tapering \citep{Furrer2006}, inducing variables \citep{Quinonero2005,Schwaighofer2003}, predictive processes \citep{Banerjee2008}, and multiresolution models \citep{Sang2011,Nychka2015}, to name just a few. 
Here, we focus mainly on the low rank approximations.

Many existing GP models assume certain smoothness properties which can be counterproductive when it comes to representing abrupt local changes. Although some less smooth kernel choices can be helpful at times, they assume stationary processes that do not adapt well to varying levels of smoothness. The undesirable smoothness characteristic of the traditional GPs could further get pronounced in approximate GP methods in general and rank-reduced approximations in particular \citep{Stein2014}. A way to overcome the limitations of low rank approximations is to recognize that the long-range dependencies tend to be of lower rank when compared to short-range dependencies. This idea has previously been explored in the context of hierarchical matrices \citep{Hackbusch2000,Bebendorf2016,Ambikasaran2016} and in multiresolution models \citep{Sang2011,Nychka2015,Katzfuss2017}.

Multiresolution GPs, seen as hierarchical models, connect collections of smooth GPs, each of which is defined over an element of a random nested partition \citep{Gramacy2008,Fox2012,Yi2017}. The long-range dependencies are captured by the GP at the top of hierarchy while the bottom-level GPs capture the local changes.
We can also view the multiresolution GPs as a hierarchical application of predictive processes---approximations of the true process arising from conditioning the initial process on parts of the data \citep{Banerjee2008,Quinonero2005}. The use of such models has recently been exploited in spatial statistics \citep{Sang2011,Nychka2015,Katzfuss2017} for modeling large spatial datasets. Refer to \citep{Fox2012} and \citep{Katzfuss2017} for overviews of these applications.

The existing multiresolution models are based on predictive processes and event though they are efficient in terms of computational complexity, they do assume full independence across the different resolutions. This independence assumption results in models which are inherently susceptible to the chosen resolution and approximations which are non-smooth at the boundaries. The latter problem stems from the fact that the multiresolution framework, e.g., \citep{Katzfuss2017}, recursively split each region at each resolution into a set of subregions. 
As discussed by \citet{Katzfuss2017b}, since the remainder process is assumed to be independent between these subregions, which can give rise to discontinuities at the region boundaries. A heuristic solution based on tapering functions is proposed in \citep{Katzfuss2017b} which employs Kanter's function as the modulating function to address this limitation. The sensitivity to the chosen resolution is partly due to the nature of the remainder process and the unconstrained representative flexibility of the GPs which manifests itself most noticeably at higher resolutions. As the size of the region under consideration decreases when the resolution increases, the remainder process may inevitably include certain aspects of data which might not be the patterns of interest. When all GPs are forced to be independent, there is no natural mechanism to constrain the representative flexibility of the GPs. 

These limitations can be addressed naturally by allowing the uncertainty to propagate across the different resolutions. We achieve this by conditioning the GPs on each other. 
Thus, here, we propose a new model which unlike the previous models that impose full independence among resolutions, instead assumes \emph{conditional independence}. Relaxing the full independence assumption is shown to result in models that are robust to overfitting in the sense of reduced sensitivity to the chosen resolution---that is regardless of the extra computational complexity, arbitrary increasing the resolution only has a small effect on the optimal model performance. Furthermore, it results in predictions which are smooth at the boundaries. 
This is facilitated by constructing a low-rank representation of the GP via a Karhunen-Lo\`eve expansion with the  Bingham prior model that consists of basis axes and basis-axis scales. Our multiresolution model ties all GPs, across all resolutions, to the same set of basis axes. These axes are learned successively in a Bayesian recursive fashion. 
We consider a fully Bayesian treatment of the proposed model and derive a structured variational inference based on a partially factorized mean-field approximation\footnote{An implementation of the model is available at: https://github.com/jtaghia/ciMRGP}.

The idea of using conditional independence in the context of multiresolution GPs has previously been studied by  \citet{Fox2012}. The two models differ in their underlying generative models and in their inference. While the computational complexity of the proposed model scales linearly with respect to the number of samples, Fox \& Dunson's model scales cubically and relies on \small$\mathrm{MCMC}$ \normalsize inference which may further limit its application to large datasets. 

Our main \textit{contribution} is to develop the conditionally independent multiresolution GP model and to derive a variational inference method to learn this model from data. 
The Bingham distribution \citep{Bingham1974} is an important distribution in directional statistics \citep{Mardia2009} where it is commonly used for shape analysis where the inference is typically based on \small{$\mathrm{MLE}$} \normalsize \citep{Kent1994b}, \small{$\mathrm{MAP}$} \normalsize \citep{Micheas2006}, and \small{$\mathrm{MCMC}$} \normalsize \citep{Leu2014}.
Hence, our use of the Bingham distribution and the corresponding variational inference solution for this model might also appeal to  researchers in directional statistics. 

\vspace{-1ex}
\section{KARHUNEN-LO\`EVE REPRESENTATION OF THE GP}
\vspace{-1ex}
\label{sec:back}
Consider a minimalistic model of GP regression,
${\vc y_t = \vc f(\vc x_t) + \vc b + \vc e_t,  \forall t\in \mathcal{T}=\{1,\ldots,n\}}$,
where ${\vc f \sim \mathcal{GP}(\cdot)}$ denotes a zero-mean GP prior, $\vc b$ denotes a constant bias, ${\vc e_t\sim \mathcal{N}(\vc 0, \gamma^{-1}\mathrm{\bc I})}$ denotes Gaussian noise with zero mean and variance $\gamma^{-1}$, ${\vc x_t\in \mathbb{R}^{d_{ x}}}$ denotes the input variables, and ${\vc y_t\in \mathbb{R}^{d_{y}}}$ denotes the measurements, $d_x,d_{ y}\in \mathbf{N}_{\geq 1}$. The standard solution involves inversion of a Gram matrix which is an $\mathcal{O}(n^3)$ operation in general. In the following, we consider low rank representations of the GP enabled via the Karhunen-Lo\`eve expansion theorem.
\vspace{-2ex}
\paragraph{Gaussian~Model}
For a $d_x$-dimensional input variable ${\vc x_{t}}$ on the interval ${[-L_{1}, L_{1}]\times \ldots\times [-L_{d_x}, L_{d_x}] \in \mathbb{R}^{d_x}}$, the GP can be represented using the (truncated) Karhunen-Lo\`eve expansion according to \citep{Solin2014}, 
\begin{gather}
\label{eq:GP_KL1}
\setlength{\belowdisplayskip}{0pt} \setlength{\belowdisplayshortskip}{0pt}
\setlength{\abovedisplayskip}{0pt} \setlength{\abovedisplayshortskip}{0pt}
\!\!\!\!\!{\vc f(\vc x_t)\! \approx \!\sum_{i=1}^{p} \!\vc w_i \phi_i(\vc x_t, \vc \tau)}, \ \forall\  {\vc w_i \!\sim \!\mathcal{N}(\vc 0, S(\sqrt{\lambda_i(\vc \tau)})\mathrm{\bc I})}, \hspace{-1ex}
\end{gather}
where ${\vc w_i=(w_{i1},\ldots,w_{id_{ y}})^\top}$ denotes the basis vectors of the series expansion, ${\vc \tau=(\tau_1, \ldots, \tau_{d_x})^{\top}}$ denotes the basis intervals such that ${\tau_d>L_d, \forall d\in \{1, \ldots, d_x\}}$, ${\phi_i (\vc x_t, \vc \tau)}$ denotes the orthogonal eigenfunctions (basis functions) with the corresponding eigenvalues $\lambda_i(\vc \tau)$, and $S(\cdot)$ denotes the spectral density of the covariance function. Note that, unlike the minimalistic representation used by \citet{Solin2014}, we have explicitly included the basis intervals $\vc \tau$ in the representation, which are treated as random variables. Their specific values are found using maximum likelihood estimation.

To ensure that the representation satisfies the dual orthogonality requirement of the Karhunen-Lo\`eve expansion, all the basis vectors~$\vc w_i$ must be zero-mean. Normally, we would assign a zero-mean Gaussian distribution over $\vc w_i$, or alternatively we  could assign a zero-mean matrix-normal distribution over ${\vc W=(\vc w_1, \ldots, \vc w_p)}$ as was done by \citet{Svensson2017}. The choice of zero-mean Gaussian priors over the basis vectors would lead to Gaussian posteriors with \emph{non-zero} means. In our multiresolution model, as we shall see later in Sec.~\ref{sec:model}, the basis vector posterior needs to be learned in a recursive fashion such that the posterior from the current resolution is used as the prior for the resolution in the next level of the hierarchy. Now, as the expansion requires the prior to be zero-mean, we would then need a posterior over basis vectors which is zero-mean by construction. If we were going to use Gaussian priors, the result would be a multiresolution model where all GPs must be fully independent. 

To address this issue, we now separate the basis vectors into two parts: \emph{basis axes} and \emph{basis-axis scales}. The basis axis vectors are defined to be \emph{antipodally symmetric}---meaning that for a random variable $\vc \vartheta$, ${p(\vc \vartheta) = p(-\vc \vartheta)}$---and thus zero-mean by construction. They primarily carry information about the direction and we can for that reason without loss of generality assume them to be on the unit sphere. The axes will be shared across resolutions such that given the axes, all GPs are independent. Although the GPs are tied to the same set of axes, they will be scaled by resolution-specific variables, namely the basis-axis scales. The 
axial distributions from directional statistics \citep{Mardia2009} make for a perfect fit in modeling these axes. In the following we consider a very specific choice of prior model, namely the \emph{Bingham distribution}, since it conveniently allows for the design of a  conditionally independent multiresolution model.
\vspace{-2ex}
\paragraph{Bingham Model}
Let $\mathcal{S}^{d-1}= \{\vc z\in \mathbb{R}^{d}: \vc z^\top \vc z=1, d\in\mathbb{N}_{>1}\}$ denote the unit  sphere. Furthermore, let ${\vc w_i\colonequals a_i \vc u_i}$ such that ${\vc u_i=\frac{\vc w_i}{\|\vc w_i\|} \in \mathcal{S}^{d_{y}-1}}$ and ${a_i = \|\vc w_i\|}$ denote the basis axes and the basis-axis scales, respectively. Without loss of generality, we can now express the noisy measurements in~\eqref{eq:GP_KL1} as 
\begin{equation}
\setlength{\belowdisplayskip}{0pt} \setlength{\belowdisplayshortskip}{0pt}
\setlength{\abovedisplayskip}{0pt} \setlength{\abovedisplayshortskip}{0pt}
  \label{eq:y3}
  \vc y_t = \sum_{i=1}^p a_i \vc u_i \phi_i(\vc x_t, \vc \tau) + \vc b + \vc e_t, 
  \qquad \forall t\in \mathcal{T}.  
\end{equation}
The basis axes ${\vc U=(\vc u_1, \ldots, \vc u_p)}$ are modeled as \emph{Bingham distributions}  \citep{Bingham1974} according to 
\begin{align*}
{p(\vc U) = \prod_{i=1}^p p(\vc u_i)}, \qquad {\forall  \ \vc u_i \sim \mathcal{B}(\bc B_i)},
\end{align*}
where $\mathcal{B}(\bc B_i)$ denotes the Bingham distribution parameterized with a real-symmetric matrix~${\bc B_i}$---the matrix $\bc B_i$ is often presented using the notion of an eigendecomposition as: ${\bc B_i=\bc M_i \times \mathrm{diag}[\vc \kappa_i]\times {\bc M_i}^\top}$ with $\bc M_i$ and $\vc \kappa_i$ being the eigenvectors and the eigenvalues of the decomposition. It is straightforward to show that $\vc u_i$ satisfies the Karhunen-Lo\`eve expansion requirements. Importantly, the Bingham distribution is antipodally symmetric, which in turn implies that ${\mathrm{E}[\vc u_i]=0}$ by construction \citep[Ch.~9.4]{Mardia2009}. We can then assign zero-mean Gaussian distributions as priors over the basis-axis scale variables~$\{a_i\}_{i=1}^p$. Assuming ${\vc e_t\sim \mathcal{N}(0, \gamma^{-1}\mathrm{\bc I})}$, and using ${\|\vc u_i \|=1}$, this choice of prior over $\vc u_i$ and $a_i$ is conveniently \emph{conjugate} to the data likelihood. 

The main constraint enforced by our choice of the Bingham prior model is the implicit requirement of ${d_y>1}$, as the Bingham density is defined on $\mathcal{S}^{d_y-1}$. For the case of ${d_y=1}$, if we assume ${\vc u_i=1}$, the Bingham model reduces to a multiresolution architecture with fully independent GPs. Other prior models should be considered for the special case of ${d_y=1}$. 
One possible choice is provided by the one-parameter version of the Bingham model \citep{Kelker1982} for modeling axes concentrated asymmetrically near a small circle. As the objective of this work is to show the advantage of the conditional independence over the full independence, we restrict our theoretical discussion to the Bingham prior model and cases where ${d_y>1}$. 
\vspace*{-1ex}
\section{MODEL}
\begin{figure*}[t]
\centering
\vspace*{-3ex}
\includegraphics[width=.99\textwidth]{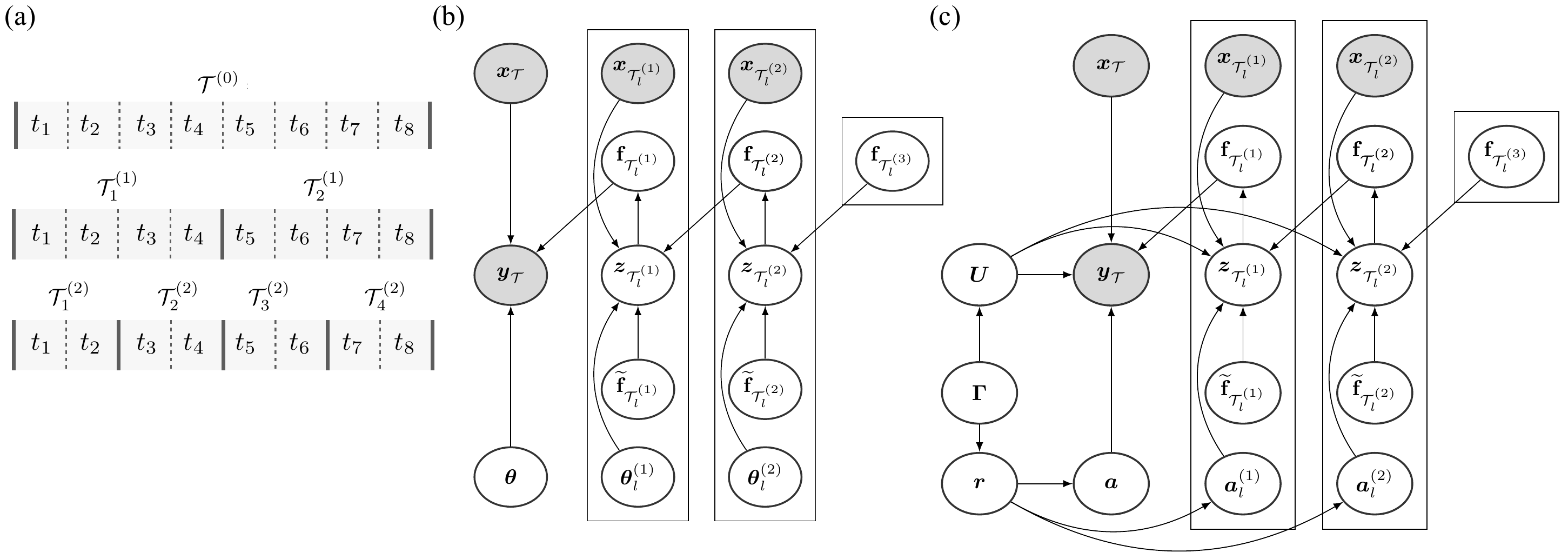}
\caption{(a) Recursive partitioning of the index set by a factor of ${2}$ for a model with resolution $m=2$. (b) The graphical representation of the fully independent MRGP (fiMRGP) model using the conventional plate notation. The boxes indicate $|\TT \h j|$ replications and the arrows show the dependency between variables. (c) The graphical representation of the conditionally independent MRGP (ciMRGP) model. Note that, for better readability, we have not shown noise and bias variables as indicated in~\eqref{eq:prior_explicit}.}
\label{fig:model}
\vspace{-2ex}
\end{figure*}

\vspace{-1ex}
\label{sec:model}
\paragraph{Notation}
Consider a recursive partitioning of the index set ${\TT=\{1,\ldots,n\}}$ across ${m}$ resolutions. At each resolution ${j\in \{1,\ldots,m\}}$, $\TT$ is partitioned into a number of non-overlapping regions. The partitioning of $\TT$ can be structured or random. Without loss of generality, consider a uniform subdivision of the index set across resolutions by a factor of $\mathfrak{q}$, such that $\TT$ is first partitioned into $\mathfrak{q}$ regions, each of which is then partitioned into ${\mathfrak{q}}$ subregions. 
The partitioning continues until resolution~$m$ where the index sets at various resolution are denoted by $\TT^{(0)} \colonequals\TT$, $\TT^{(1)}=\{ \TT^{(0)}_1, \ldots, \TT^{(0)}_{\mathfrak{q}} \}$, and similarly by $\TT^{(m)}=\{ \TT^{(m-1)}_1, \ldots, \TT^{(m-1)}_{\mathfrak{q}} \}$,
where ${ | \TT^{(0)} |= 1}$, ${| \TT^{(1)} |= {\mathfrak{q}}}$, and ${| \TT^{(m)} |= {\mathfrak{q}}^m}$.
An example of such a partitioning by a factor of ${\mathfrak{q}=2}$ is shown in Fig.~\ref{fig:model}-a. 
As a convention, we will use the notation $\TT_l^{(j)}$ to indicate the $l$-th element of the set ${\TT^{(j)}=\{\TT_l^{(j)}\}_{l=1}^{| \TT^{(j)} |}}$, which corresponds to the index set related to region $l$ at resolution~$j$. We also define ${\vc x_{\TT^{(0)}}\colonequals\xt}$ and ${\vc x_{\TT^{(j)}} = \{\vc x_{\TT_l^{(j)}}\}_{l=1}^{|\TT\h j|}}$, where
${\vc x_{\TT_l^{(j)}}=\{\vc x_t \mid \forall t\in \TT_l^{(j)}\}}$.

\vspace*{-1ex}
 \paragraph{Generative Model}
As before, let $\vc f (\cdot )$ be the stochastic process of interest. Once the process is observed at $\vc x_{\TT}$, it gives rise to the noisy observations~$\vc y_t$. By making use of a Gaussian process as the prior over $\vc f(\cdot)$, the observations ${\vc y_t}$ at resolution ${j=0}$ are modeled according to~\eqref{eq:y3}.
 In a multiresolution setting based on the hierarchical application of predictive processes, we approximate $\vc f(\cdot)$ according to 
\begin{align*}
{\vc f (\cdot ) = \widehat{\vc f}~\!\!\h 0 (\cdot) + \vc f~\!\!\h 1(\cdot )}, 
\end{align*}
where ${\widehat{\vc f}~\!\!\h 0}$ is the approximate predictive process at resolution ${j=0}$, and $\vc f~\!\! \h 1(\cdot ) $ is the so-called \emph{remainder process}. Let $\vc z_{t,l}\h 1$ indicate the noisy instantiations of the latent process $\vc f~\!\!\h 1(\cdot ) $ at $\vc x_{\TT\h 1}$. We will treat $\vc z_{t,l}\h 1$ as a \emph{latent variable}, and model it using a conditionally independent GP prior, for all ${\vc x_t \in \vc x_{\TT_l^{(1)}}}$,
\begin{align*}
\setlength{\belowdisplayskip}{0pt} \setlength{\belowdisplayshortskip}{0pt}
\setlength{\abovedisplayskip}{0pt} \setlength{\abovedisplayshortskip}{0pt}
{\vc z_{t,l}^{(1)} = \sum_{i=1}^p a_{i,l}^{(1)} \vc u_i \phi_{i}\h 1(\vc x_t, \vc \tau_l \h 1) + \vc b_l\h 1 + 
\vc e_{t,l}^{(1)}}, 
\end{align*}
where the basis axes~$\vc u_i$ are shared among all the processes while the basis-axis scales~$a_{i,l}^{(1)}$ are region specific. At the higher resolution, ${j = 2}$, the latent process $\vc f \h 1(\cdot ) $ is in turn approximated by ${\vc f~\!\!\h 1 (\cdot ) = \widehat{\vc f}~\!\!\h 1 (\cdot) + \vc f ~\!\!\h 2(\cdot )}$. In general, for resolution~$j$ we have
\begin{align*}
  {\vc f~\!\!\h j (\cdot ) = \widehat{\vc f}~\!\!\h j (\cdot) + \vc f~\!\! \h {j+1}(\cdot )},
\end{align*}
where $\vc f \h {j+1}(\cdot )$ is the remainder process at resolution ${j+1}$ whose noisy instantiations on $\TT^{(j+1)}$ are modeled according to, ${\forall \vc x_t \in \vc x_{\TT_l^{(j+1)}}}$:
\begin{align*}
{\vc z_{t,l}^{(j+1)} \!=\! \sum_{i=1}^p \!a_{i,l}^{(j+1)} \vc u_i \phi_{i}\h{j+1}(\vc x_t, \vc \tau_l\h {j+1}) \!+ \!\vc b_l\h {j+1}\! + \!\vc e_{t,l}^{(j+1)}}.
\end{align*}
Throughout, $\vc u_i$ has been written without indexing w.r.t.~$l$ and~$j$. This is to emphasize that these are shared across all resolutions and regions such that in transition from one resolution to another, the axes of the basis vectors remain the same but they may be scaled differently via a region-specific and resolution-specific variable $a_{i,l}^{(j)}$. 
The noise variable is indexed w.r.t. both~$l$ and~$j$, but we could alternatively assume the noise to be a resolution-specific variable. 
In a multiresolution model, bias may not be simply removed as a part of the preprocessing step, as the bias at each resolution carries uncertainties from the previous resolutions.
 These parameters are expressed using indexing on both $j$ and $l$. 
We have indicated the basis functions with indexing on $j$, as generally one might consider a different choice of basis functions at different resolutions. The basis interval variables $\vc \tau_l\h j$ are learned from data and expressed with both $j$ and $l$.

The recursive procedure continues until resolution ${j=m}$ is reached. By assuming that the latent remainder process at ${j=m+1}$ approaches zero, we can approximate~$\vc f(\cdot)$ as the sum of the predictive processes from all resolutions,
\begin{align*}
{\vc f(\cdot) = \vc f~\!\!\h {m+1}(\cdot)+\sum_{j=0}^m \widehat{\vc f}~\!\!\h j(\cdot) 
\approx \sum_{j=0}^m \widehat{\vc f}~\!\!\h j(\cdot)},
\end{align*}
where $\widehat{\vc f}~\!\!\h 0$ captures global patterns and finer details are captured at higher resolutions. 
\vspace*{-1ex}
\section{BAYESIAN INFERENCE}
\label{sec:bayesian_learning}
\vspace*{-1ex}
\paragraph{Notation}
Let ${\vc y_{\TT^{(0)}}\colonequals\yt}$ where ${\yt=\left\{\vc y_{t} \mid  \forall t\in \TT\right\}}$ denote the set of noisy observations, and ${\vc z_{\TT \h j}= \{\vc z_{\TT_l \h j}\}_{l=1}^{|\TT \h j|}}$  denote the set of latent variables for ${j\geq 1}$, where
${\vc z_{\TT_l^{(j)}}=\{\vc z_{t,l}^{(j)} \mid  \forall t\in \TT_l^{(j)}\}}$.
We denote the latent function instantiations at $\vc x_{\TT_l\h j}$ by ${\bc f_{\TT_l \h j}=\{\bc f\h j_l(\vc x_{t})\equiv \bc f\h j_{l,t}\mid \forall \vc x_t \in \vc x_{\TT_l \h j}\}}$. Similarly, let ${\bc f_{\TT\h j}=\{\bc f_{\TT_l\h j}\}_{l=1}^{| \TT\h j |}}$.
Furthermore, to keep the notation uncluttered, let: 
\begin{align*}
&{\underline{\zt}=\{\ztj j\}_{j=1}^{m}}, \\ &  {\underline{\xt}=\{\vc x_{\TT \h j}\}_{j=0}^{m}},
\\
&
 {\underline{\bc f_{\TT}}= \{\bc f_{\TT\h j}\}_{j=0}^{m}}, \quad
\underline{\widetilde{\bc f}}\h j =  \{\ftj {j^{\prime}}\}_{j^{\prime}=0}^{j-1}, \forall j\geq 1,
\\
&{\underline{\vc a}=\left\{\{\vc a_l\h j\}_{l=1}^{|\TT\h j|}\right\}_{j=0}^{m}}, \ \vc a\h 0\equiv \vc a, \\
& {\underline{\gamma}=\left\{\{\gamma_l\h j\}_{l=1}^{|\TT\h j|}\right\}_{j=0}^{m}}, \ \gamma\h 0\equiv \gamma, \\
&{\underline{\vc b}=\left\{\{\vc b_l\h j\}_{l=1}^{|\TT\h j|}\right\}_{j=0}^{m}}, \ \vc b\h 0\equiv \vc b, \\ & \underline{\vc \theta} = \left\{ \{\vc \theta_l \h j \}_{l=1}^{|\TT\h j|}\right \}_{j=0}^{m},\  {\vc \theta_l \h j \!= \!\{\vc a_l \h j\!, \vc U\!,\vc b_l\h j\!,\! \gamma_l\h j\}}.
\end{align*}
We first discuss the design of a fully independent model and its limitation. We then introduce the case of the conditionally independent model.
\subsection{Fully Independent MRGP}
\paragraph{Joint Distribution} 
The joint distribution of all observations and all latent variables is expressed as
\begin{multline}
\label{eq:joint_ind}
\setlength{\belowdisplayskip}{0pt} \setlength{\belowdisplayshortskip}{0pt}
\setlength{\abovedisplayskip}{0pt} \setlength{\abovedisplayshortskip}{0pt}
p\big(\yt , \underline{\zt}, \underline{\bc f_{\TT}}, \ftj {m+1}, \underline{\xt},\underline{\vc \theta}\big) \\ =
p(\yt \!\mid\! \ftjl {1}l , \xt, \vc \theta\h0) p(\vc \theta\h0) \\
 \times \Bigg[\! \prod_{j=1}^m \!\!\prod_{l=1}^{|\TT^{(j)}|} \! p(\ztjl jl  \!\mid\! \ftjl {j+1}l, \underline{\widetilde{\bc f}}_l\h j \!, \xtjl jl, \vc \theta_l \h j)p(\vc \theta_l \h j)\Bigg] \\ \times \Bigg[\prod_{j=1}^{m} \prod_{l=1}^{|\TT^{(j)}|}  p(\ftjl {j}l \!\mid \!\ztjl {j}l) \Bigg] p(\ftj {m+1}).
\end{multline}
The corresponding graphical representation of the model is shown in Fig~\ref{fig:model}-b, for the special case of ${m=2}$.
\paragraph{Variational Inference} Using variational inference \citep{Jordan1999,BleiKM:2017}, the goal is to find a tractable approximation of the true posterior distribution. Consider a variational posterior in the form of:
\begin{multline}
\label{eq:post_in}
\setlength{\belowdisplayskip}{0pt} \setlength{\belowdisplayshortskip}{0pt}
\setlength{\abovedisplayskip}{0pt} \setlength{\abovedisplayshortskip}{0pt}
\! \!q(\underline{\zt}, \underline{\bc f_{\TT}}, \ftj {m+1}, \underline{\vc \theta}) =
\Bigg[\prod_{j=0}^{m} \prod_{l=1}^{|\TT^{(j)}|} q(\vc \theta_l\h j )\Bigg] \\ \times
\Bigg[\prod_{j=1}^{m} \!\prod_{l=1}^{|\TT^{(j)}|} \!\! q(\ztjl jl) q(\ftjl {j}l \!\mid \!\ztjl {j}l) 
  \Bigg] q(\ftj {m+1}). \hspace{-1ex}
\end{multline}
Using the mean-field assumption and choosing conjugate priors, it is possible to find tractable expressions for $q(\vc \theta_l\h j )$ and $q(\ztjl jl)$. However, $q(\ftj {m+1})$ and $q(\ftjl {j}l \!\mid \!\ztjl {j}l)$ can still be intractable. Following a similar approach as in \citep{Frigola2014} and \citep{DamianouL2013}, we can take $q(\ftj {m+1})$ and ${q(\ftjl {j}l \!\mid \!\ztjl {j}l) }$ to match the prior model. These difficult-to-compute terms would then effectively cancel in the optimization when computing the Kullback-Leibler divergence between the prior and posterior. This simplifying assumption, in particular for ${q(\ftjl {j}l \!\mid \!\ztjl {j}l)}$, makes the inference tractable but it comes with the price of severely underestimating uncertainties which ultimately causes overfitting in terms of sensitivity to the chosen resolution. 

To reduce the implications of this simplification while maintaining a tractable solution, we will allow the GPs to share part of the parameter space $\underline{\vc \theta}$. In the following, we discuss this model alternative.
\vspace*{-2ex}
\subsection{Conditionally Independent MRGP}
\vspace*{-2ex}
\paragraph{Joint Distribution}
The joint distribution of all observations and all latent variables is given by
\begin{multline}
\label{eq:joint}
p\big(\yt , \underline{\zt}, \underline{\bc f_{\TT}}, \ftj {m+1}, \underline{\xt},{\vc U, \underline{\vc a}, \underline{\vc b}, \underline{\gamma}}, \vc \Gamma,\vc r\big) \\ =
p(\yt \!\mid\! \ftjl {1}l , \xt, \vc \theta\h0) p(\vc \theta\h0 \!\mid \!\vc \Gamma, \vc r) \\
 \times \Bigg[\! \prod_{j=1}^m \!\!\prod_{l=1}^{|\TT^{(j)}|} \! p(\ztjl jl  \!\mid\! \ftjl {j+1}l, \underline{\widetilde{\bc f}}_l\h j \!, \xtjl jl, \vc \theta_l \h j)p(\vc \theta_l \h j, \vc \Gamma,\! \vc r)\Bigg] \\ \times \Bigg[\prod_{j=1}^{m} \prod_{l=1}^{|\TT^{(j)}|}  p(\ftjl {j}l \!\mid \!\ztjl {j}l) \Bigg] p(\ftj {m+1}), 
\end{multline}
where 
the pair of $\vc \Gamma$ and $\vc r$ are hierarchical parameters which will be discussed shortly. The corresponding graphical model is shown in Fig.~\ref{fig:model}-c.

The prior model parameter in~\eqref{eq:joint} is factorized as
\begin{multline}
\label{eq:prior_explicit}
p(\vc \theta_l \h j, \vc \Gamma, \vc r) = p(\vc b_l \h j \mid \gamma_l \h j) p(\gamma_l \h j) p(\vc U\mid \vc \Gamma)  \\  \times p(\vc a\h j_l \mid \!\vc r)  p(\vc r \mid  \vc \Gamma) p(\vc \Gamma).
\end{multline}
To facilitate expressions of the conditional distributions, let $\mathfrak{Z}_{k}\h j, \forall k\in\{j, j+1\}$, indicate a binary switch parameter such that ${\mathfrak{Z}_{k}\h j = 1}$ when ${k=j}$ and ${\mathfrak{Z}_{k}\h j = 0}$ when ${k=j+1}$.
The conditional distribution of the observations is expressed by 
\begin{multline*}
p(\yt \mid \ftj {1}, \xt, \vc \theta) = \prod_{k\in \{0, 1\}}\!\!\Bigg[\!\prod_{l=1}^{|\TT\h {1}|}\!\!\!\prod_{t\in\TT_l^{(1)}} \!\!\!\!p_0(\bc f_t)\!\Bigg]^{1-\mathfrak{Z}_k\h j} \\ 
\times \Bigg[\!\prod_{t\in \TT\h 0} \!\! \!\mathcal{N}\big(\vc y_{t}; \vc b \!+ \!\sum_{i=1}^p\! a_{i}\vc u_i \phi_i\h 0(\vc x_t, \!\vc \tau\h 0) , {\gamma}^{-1} \big) \Bigg]^{\mathfrak{Z}_k \h j}\!\!\!\!\!, 
\end{multline*}
and the
conditional distribution of the latent variables $\ztjl jl$, $\forall j$, is expressed by
\begin{align*}
p(\ztjl jl  \mid \ftj {j+1}, \underline{\widetilde{\bc f}}_l\h j, \xtjl jl, \vc \theta_l\h j) ~~~~~~~~~~~~~~~~~~~~~~~~  \\=  \prod_{k\in \{j, j+1\}} \Bigg[\prod_{l=1}^{|\TT\h {j+1}|}\prod_{t\in\TT_l^{(j+1)}}p_0(\bc f_t)\Bigg]^{1-\mathfrak{Z}_k\h j} \\ \times  \Bigg[\prod_{t\in \TT_l^{(j)}} \mathcal{N}\big(\vc z_{t,l}\h {j}; \bar {\vc z}_{t,l}\h j , {\gamma_l\h j}^{-1} \big) \Bigg]^{\mathfrak{Z}_k\h j},
\end{align*}
where ${\bar {\vc z}_{t,l}\h j}$, $\forall j\geq 1$, is defined as:
\begin{align*}
{\bar {\vc z}_{t,l}\h j = \sum_{j^{\prime}=0}^{j-1} \bc f_{t,l}^{j^{\prime}} + \vc b_l \h j + \sum_{i=1}^p a_{i,l}\h j \vc u_i \phi_i\h j(\vc x_t, \vc \tau_l\h j)},
\end{align*}
 and $p_0(\bc f_t)$ approaches the Dirac point mass $\delta(\bc f_t)$.
\vspace{-1ex}
\paragraph*{Role of Hierarchical Parameters}
As mentioned earlier, in the expression for the joint distribution~\eqref{eq:joint} we have introduced hierarchical parameters ${\vc \Gamma=[\Gamma_{ik}], i,k\in \{1, \ldots, p\}}$ and ${\vc r\!=\!(r_1, \ldots,r_p)^{\top}}$, which are not explicit in the generative model, Fig.~\ref{fig:model}-b.

 The parameters $r_{i}$ represent the precision of the basis-axis scale parameter $a_{i,l}^{(j)}$ and are shared across resolutions and regions. These parameters will enable automatic determination of the effective number of basis axes, as the posterior will approach zero for axes that are effectively not used. Thus at each resolution and in each region, only a subset of the basis axes will be used and others will have little to no influence. 

Furthermore, our recursive framework requires the indexing of the axes of $\vc U$ to be the same across resolutions. More precisely, we shall learn the posterior distribution over $\vc U$ in a Bayesian recursive fashion such that the posterior from the previous resolution is used as the prior for the current resolution. A complication is that the indexing of ${\{\vc u_i\}_{i=1}^{p}}$ might end up being completely arbitrary at each resolution. This is because $\vc u_i$ is distributed according to a Bingham distribution as ${\vc u_i\sim \mathcal{B}(\bc B_i)}$, where ${\bc B_i}$ is expressed via a set of eigenvectors and eigenvalues, ${\bc B_i=\bc M_i \times \mathrm{diag}[\vc \kappa_i]\times {\bc M_i}^\top}$. 
The complication is that the indexing of these eigenvectors can be completely arbitrary, implying that the necessary one-to-one correspondence between the eigenvectors representing the prior and those representing the posterior is lost. 
Our sequential (recursive) learning however requires a unique one-to-one correspondence. We might consider to sort the eigenvectors (axes) based on their corresponding eigenvalues. However, that would result in sub-optimal performance.

 To formally handle the axis-index ambiguity across resolutions, we have introduced a latent sparse matrix~$\vc \Gamma$ of binary indicator variables to account for the possible index permutation between the prior and the posterior of the basis axes 
in transitioning from resolution ${j-1}$ to $j$. A matrix element ${\Gamma_{ik}=1}$ indicates that the axis identified by index $k$ in the posterior model of resolution ${j-1}$ is identical to the axis denoted by index $i$ in the current resolution $j$. In defining the prior, Eq.~\eqref{eq:pu} and Eq.~\eqref{eq:prior_lambda}, we have conditioned both $\vc u_i$ and $\vc r$ on $\vc \Gamma$ to ensure accumulation of ``aligned prior beliefs'' of these parameters across resolutions (see~\eqref{eq:prior_explicit} and Fig.~\ref{fig:model}-c).

The explicit form of the prior distributions over all variables in~\eqref{eq:prior_explicit} is discussed in detail in App.~\ref{app:prior}. 
\vspace*{-2ex}
\paragraph{Variational Inference}
\label{sec:vi}
Here, we consider a variational posterior in the form of:
\label{eq:post}
\begin{multline*}
\! \!q(\underline{\zt}, \underline{\bc f_{\TT}}, \ftj {m+1},\vc U, \underline{\vc a},\vc \Gamma,{\vc r}) \!\!=\!\!
\Bigg[\!\prod_{j=0}^{m} \!\!\prod_{l=1}^{|\TT^{(j)}|}\!\!q(\vc \theta_l\h j\!, \vc \Gamma,\! \vc r)\Bigg] \\ \times
\Bigg[\prod_{j=1}^{m} \!\prod_{l=1}^{|\TT^{(j)}|} \!\! q(\ztjl jl)  p(\ftjl {j}l \!\mid \!\ztjl {j}l) 
  \Bigg] p(\ftj {m+1}), \hspace{-1ex}
\end{multline*}
where the use of a \emph{partially factorized} mean-field approximation results in
\begin{multline}
\label{eq:post_explicit}
q(\vc \theta_l\h j, \vc \Gamma, \vc r)  = q(\vc b_l \h j \mid \gamma_l\h j) q(\gamma_l\h j) \\ \times q(\vc a\h j_l \!\mid\! \vc U)q(\vc U) q( \vc r) q(\vc \Gamma). 
\end{multline}
We then take $p(\ftj {m+1})$ and ${p(\ftj j\mid \ytj j)}$ to match the ones in the prior model of the joint expression~\eqref{eq:joint} allowing a tractable solution. Furthermore, notice the difference in factorization of the prior~\eqref{eq:prior_explicit} and the posterior~\eqref{eq:post_explicit}. In particular, we have considered a joint posterior over basis axes and their scales, ${q(\vc u_i) q(a_{i,l}^{(j)} \!\mid \vc u_i) }$. The joint posterior allows us to conveniently use the posterior $q(\vc u_i)$ as the prior in the factorized prior for the sequential (recursive) learning procedure.

Given the joint distribution and our choice of the variational posterior distribution, the variational lower bound is expressed by
\begin{align}
\label{eq:LL}
\mathcal{L}= \mathcal{L}_{\yt} + \sum_{j=1}^{m} \mathcal{L}_{\ztj j},
\end{align}
where $\mathcal{L}_{\yt}$ can be written as the sum of the likelihood and the negative Kullback-Leibler divergence (KLD) between the posterior and the prior, 
\begin{multline*}
\mathcal{L}_{\yt} =  \la \log p(\yt  \!\mid \!\xt, \ftj 1, \vc \theta\h 0) \ra_{ q(\vc \theta\h 0)p(\ftj 1)} \\ - \la \log \frac{q(\vc \theta\h 0, \vc \Gamma,\vc r )}{p(\vc \theta\h 0, \vc \Gamma,\vc r )} \ra_{q(\vc\theta\h 0, \vc \Gamma,\vc r )}.
\end{multline*}
The notation $\la \cdot \ra_{q(\cdot)}$ is used to denote the expectation with respect to its variational posterior distribution.
Similarly ${\mathcal{L}_{\ztj j}}$ can be expressed as the sum of the likelihood and the negative KLD between the posterior and the prior plus the posterior entropy of the remainder term,
\begin{multline*}
\!\!\mathcal{L}_{\ztj j} \!= \! \la\! \log p(\ztj j \mid \xtj j, \ftj {j+1}, \underline{\widetilde{\bc f}}\h j, \vc \theta\h j)\! \ra_{q(\cdot)p(\cdot)} \\ - \la \log \frac{q(\vc \theta\h j,\vc \Gamma,\vc r )}{p(\vc \theta\h j, \vc \Gamma,\vc r )} \ra_{q(\vc \theta\h j, \vc \Gamma,\vc r )} - \la \log q(\ztj j) \ra_{q(\ztj j)},
\end{multline*}
where ${q(\cdot)p(\cdot) := q(\ztj {j})q(\vc \theta\h j\!)p(\underline{\widetilde{\bc f}}\h j)p(\ftj {j+1})}$.
Taking into account the convenient form of~\eqref{eq:LL}, the optimal posterior distribution can now be obtained by maximizing the lower bound using standard variational inference.

The explicit forms of the optimized variational posterior distributions are derived in App.~\ref{app:post}. 
Descriptive statistics of the posterior distributions are summarized in App.~\ref{app:stats}. The predictive process is discussed in App.~\ref{app:pred}. The optimization of the basis interval parameters is discussed in App.~\ref{app:tau}. 
Finally, an algorithmic presentation of the model is described in App.~\ref{app:alg}.
\begin{figure}[t]
\vspace*{-2ex}
\begin{center}
\centerline{\includegraphics[width=0.993\columnwidth]{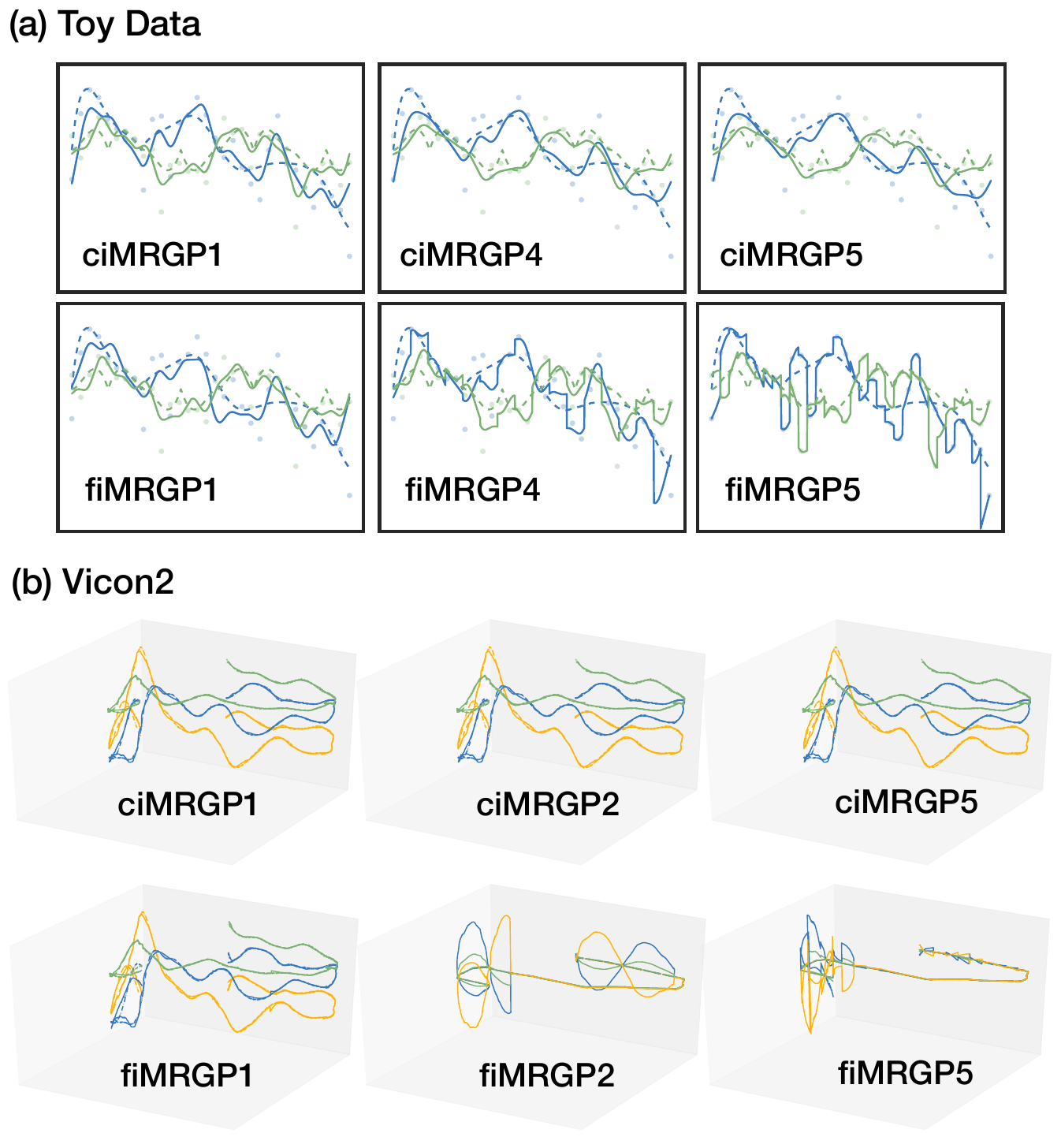}}
\vspace{-1ex}
\caption{Illustrative comparison of $\mathrm{ciMRGP}$ and $\mathrm{fiMRGP}$ at various resolutions on (a) the synthetic dataset $\mathrm{ToyData}$, App.~\ref{app:toy_data}, and (b) the real dataset $\mathrm{vicon2}$, App.~\ref{app:vicon2}. See the text for details.}
\label{fig:overfit}
\end{center}
\vskip -0.3in
\end{figure}
\begin{table*}[t!]
\vspace*{-3ex}
\footnotesize
\caption{Summary of datasets and methods used in the comparison.}
\vspace*{-3ex}
\begin{center}
\label{tb:data}
\vline
\begin{tabular}{lllllll} 
\hline
\multicolumn{2}{c}{Dataset} \\
\cline{1-2}
Name    & Source & $d_x$  & $d_y$ & $n_{\mathrm{train}}$ & $n_{\mathrm{test}}$ & Note \\
\hline
$\mathrm{ oes10}$    & \citep{Spyromitros-Xioufis2016}   & $298$  & $16$  & $302$ & $100$ & \ref{app:oes}\\
$\mathrm{ oes97}$    & \citep{Spyromitros-Xioufis2016}   & $263$  & $16$  & $250$ & $83$ & \ref{app:oes}\\
$\mathrm{ atp1d}$    &  \citep{Spyromitros-Xioufis2016}  & $411$  & $6$  & $303$ & $33$ & \ref{app:atp7d} \\
$\mathrm{ atp7d}$    &  \citep{Spyromitros-Xioufis2016}  & $411$  & $6$  & $221$ & $74$ & \ref{app:atp7d}\\
$\mathrm{ scm1d}$-a    &  \citep{Spyromitros-Xioufis2016}  & $280$  & $16$  & $2249$ & $750$ & \ref{app:scm}\\
$\mathrm{ scm1d}$    &  \citep{Spyromitros-Xioufis2016}  & $280$  & $16$  & $7352$ & $2450$ & \ref{app:scm}\\
$\mathrm{ scm20d}$    &  \citep{Spyromitros-Xioufis2016}  & $61$  & $16$  & $6724$ & $2241$ & \ref{app:scm}\\
$\mathrm{ naval}$    &  \citep{Coraddu2013}  & $16$  & $2$  & $8951$ & $983$ & \ref{app:naval}\\
$\mathrm{ vicon}$    &  \citep{Jidling2017}   & $3$  & $3$  & $8806$ & $8806$ & \ref{app:vicon} \\
$\mathrm{ hrtf }$    &  \citep{algazi01}  & $8$  & $200$  & $29$ & $8$ & \ref{app:hrtf}\\
$\mathrm{ nengo }$    &  \citep{Trevor12014,Taghia2018}  & $1$  & $7$  & $1211$ & $403$ & \ref{app:nengo}\\
$\mathrm{ lorenz96}$    & \tiny synthetic\normalsize  & $1$  & $20$  & $1000$ & $10^5$ &  \ref{app:lorenz}\\
\hline
\end{tabular} \!\vline \!\!
\begin{tabular}{lll}
\hline
\multicolumn{2}{c}{Method} \\
\cline{1-2}
Name  & Source & Note \\
\hline
$\mathrm{ MRGP0}$     &  this paper    & $m=0$, $\mathfrak{q}=2$      \\
$\mathrm{ciMRGP1}$      &  this paper   & $m=1$, $\mathfrak{q}=2$       \\
$\mathrm{ciMRGP2}$      &  this paper    & $m=2$, $\mathfrak{q}=2$      \\
$\mathrm{ciMRGP3}$      &  this paper    & $m=3$, $\mathfrak{q}=2$       \\
$\mathrm{ciMRGP8}$      &  this paper    & $m=8$, $\mathfrak{q}=2$       \\
$\mathrm{fiMRGP1}$      & this paper    & $m=1$, $\mathfrak{q}=2$       \\
$\mathrm{fiMRGP2}$      &  this paper    & $m=2$, $\mathfrak{q}=2$      \\
$\mathrm{fiMRGP3}$      &  this paper   & $m=3$, $\mathfrak{q}=2$       \\
$\mathrm{SGPMC}$      & \citep{Hensman2015}    & \ref{app:sgpmc}     \\
$\mathrm{SVGP}$      &  \citep{Snelson2006}  & \ref{app:svgp}        \\
$\mathrm{SVIGP}$     &  \citep{hensman2013}   & \ref{app:svigp}  \\    \vspace*{.15ex} \\  
\hline
\end{tabular}\vline
\renewcommand{\arraystretch}{1}
\caption{Average test RMSE for all methods across five repetitions.}
\vspace*{-2ex}
\footnotesize
\setlength\tabcolsep{2.2pt}
\begin{tabular}{l*{11}{c}r}
\hline
\label{tb:regressionMSE}
Dataset              &\tiny $\mathrm{ MRGP0}$ & \tiny $\mathrm{ciMRGP1}$ & \tiny $\mathrm{ciMRGP2}$  &\tiny  $\mathrm{ciMRGP3}$ & \tiny  $\mathrm{ciMRGP8}$&\tiny $\mathrm{fiMRGP1}$  & \tiny $\mathrm{fiMRGP2}$ & \tiny $\mathrm{fiMRGP3}$  &  \tiny $\mathrm{SGPMC}$ & \tiny $\mathrm{SVGP}$ & \tiny $\mathrm{SVIGP}$  \\
\hline
 $\mathrm{ oes10}$   &  $0.784$ & $\vc{0.757}$ & $\vc{0.757}$ & $0.758$ & --- &$0.785$ & $0.788$ & $0.799$ & $0.775$ & $0.774$ & $0.775$  \\
  $\mathrm{ oes97}$   & $0.702$ & $0.699$ & $0.697$ & $\vc{0.696}$ & ---  &$0.703$ & $0.707$ & $0.720$ & $0.705$ & $0.705$ & $0.705$  \\
 $\mathrm{ atp1d}$   & $1.334$ & $1.297$ & $1.293$ & ${1.291}$ & ---  &$1.313$ & $1.312$ & $1.309$ & $\vc{1.039}$ & $\vc{1.039}$ & $\vc{1.039}$\\
$\mathrm{ atp7d}$    &  $1.228$ & $1.231$ & $1.229$ & $1.232$ & ---  &$1.226$ & $1.222$ & $1.217$ & $\vc{1.005}$ & $\vc{1.005}$ & $1.006$\\
$\mathrm{ scm1d}$-a    & $0.887$ & $0.884$ & $0.882$  & $0.880$ & $\vc{0.871}$ & $\text{large}$ & $\text{large}$ & $\text{large}$ & $0.994$  & $1.001$ & $1.002$  \\
$\mathrm{ scm1d}$    & $1.073$ & $1.052$ & $1.047$  & $1.041$ & ${1.021}$ & $\text{large}$ & $\text{large}$ & $\text{large}$ & $\vc{1.018}$  & $\vc{1.018}$ & $1.021$ \\
$\mathrm{ scm20d}$ & $1.053$ & $1.051$ & $1.048$ & $1.042$ & $\vc{0.990}$  &$\text{large}$ & $\text{large}$ & $\text{large}$ & ${0.996}$ & $0.996$ & $0.997$\\
$\mathrm{ naval}$    & $0.009$ & $0.006$  & ${0.005}$ &  $0.005$ &$0.005$ &$\vc{0.004}$ & $0.531$ & $\text{large}$ & $0.011$ & $0.011$ & $0.019$ \\
 $\mathrm{ vicon}$   &  $0.019$ & $0.018$ & ${0.018}$ &  ${0.018}$ & $\vc{0.017}$ & $0.026$ & $\text{large}$ & $\text{large}$ & $0.326$ & $0.325$ & $0.326$ \\
$\mathrm{ hrtf}$     &  $0.015$ & $\vc{0.014}$ & $\vc{0.014}$ &  $\vc{0.014}$ &--- &$0.015$ & $0.016$ & $0.019$ & $\vc{0.014}$ & $\vc{0.014}$ & $\vc{0.014}$\\ 
$\mathrm{ nengo}$    & $0.593$ & $0.574$ & $0.564$ & ${0.561}$ &$\vc{0.552}$ &$0.594$ & $0.591$ & $0.603$ & $0.813$ & $0.812$ & $0.824$ \\
$\mathrm{ lorenz96}$ &  $0.361$ & $\vc{0.329}$  & $\vc{0.329}$ & $0.330$ & $0.330$ & $0.433$ & $\text{large}$ & $\text{large}$ & $4.142$ & $4.018$ & $4.121$ \\
\hline
\end{tabular}

\renewcommand{\arraystretch}{1}
\caption{Average test MLL for all methods across five repetitions.}
\vspace*{-2ex}
\footnotesize
\setlength\tabcolsep{2.4pt}
\begin{tabular}{l*{11}{c}r}
\hline
\label{tb:regressionMLL}
Dataset              &\tiny $\mathrm{ MRGP0}$ & \tiny $\mathrm{ciMRGP1}$ & \tiny $\mathrm{ciMRGP2}$  &\tiny  $\mathrm{ciMRGP3}$ & \tiny  $\mathrm{ciMRGP8}$  &\tiny $\mathrm{fiMRGP1}$  & \tiny $\mathrm{fiMRGP2}$ & \tiny $\mathrm{fiMRGP3}$  &  \tiny $\mathrm{SGPMC}$ & \tiny $\mathrm{SVGP}$ & \tiny $\mathrm{SVIGP}$  \\
\hline
 $\mathrm{ oes10}$   & -$9.7$ & -$4.8$ & -$3.9$ & -$\vc{3.4}$ & --- & -$9.7$ & -$9.8$ & -$10.1$ & -$5.6$ & -$5.7$ & -$10.4$  \\
  $\mathrm{ oes97}$   &  -$5.0$ & -$2.9$ & -$2.6$ & -$\vc{2.5}$ & ---& -$5.0$ & -$5.1$ & -$5.3$ & -$4.8$ & -$4.8$ & -$8.7$  \\
 $\mathrm{ atp1d}$   & -$19.0$ & -$6.7$ & -$3.9$  & -$\vc{2.9}$ &--- & -$18.6$ & -$18.5$ & -$18.5$ & -$4.1$ & -$4.1$ & -$7.5$ \\
$\mathrm{ atp7d}$    &  -$11.3$ & -$4.7$ & -$3.3$ & -$\vc{2.7}$ &--- &-$11.2$ & -$11.2$ & -$11.1$ & -$3.9$ & -$3.9$ & -$7.1$\\
$\mathrm{ scm1d}$-a    & -$56.8$ & -$25.5$ & -$16.9$ & -$12.5$ &-$\vc{2.9}$ & -$\text{large}$ & -$\text{large}$ & -$\text{large}$& -${8.8}$ & -$8.8$ & -$\text{large}$ \\
$\mathrm{ scm1d}$    & -$98.2$ & -$46.3$ & -$30.8$ & -$22.8$ &-$\vc{8.0}$ & -$\text{large}$ & -$\text{large}$ & -$\text{large}$& -${9.2}$ & -$9.1$ & -$\text{large}$ \\
$\mathrm{ scm20d}$ & -$92.2$ & -$41.5$ & -$27.9$ & -$20.6$ & -$\vc{7.5}$ &-$\text{large}$ & -$\text{large}$ & -$\text{large}$& -$8.9$ & -$8.8$ & -$\text{large}$\\ 
$\mathrm{ naval}$    &  $1.8$ & $2.2$ & $2.3$ & ${3.3}$ &${3.4}$ &$\vc{3.6}$ & -$2497.$ & -$\text{large}$ & $1.2$ & -$0.9$ & -$47.2$ \\
 $\mathrm{ vicon}$   & -$0.5$ &-$1.2$ & -$0.9$ &-${0.3}$ & -$\vc{0.2}$ &-$4.8$ & -$\text{large}$ & -$\text{large}$ & -$14.5$ & -$1.0$ &-$\text{large}$\\
$\mathrm{ hrtf}$     & $\vc{0.1}$ & -$0.0$ & -$0.5$ & -$0.9$ &--- &$ 0.0$ & $0.0$ & -$0.2$ & -$0.9$ & -$0.9$ & -$0.3$\\
$\mathrm{ nengo}$    &  -$22.6$ & -$7.0$ & -$4.8$ & -${3.6}$ &-$\vc{2.3}$ &-$27.0$ & -$26.4$ & -$25.8$ & -$228.$ & -$3.2$  & -$23.4$\\
$\mathrm{ lorenz96}$ & -$139.$ & -$30.6$ & -$16.1$ & -${9.3}$ & -$\vc{3.1}$ &-$109.$ & -$\text{large}$ & $\text{large}$ & -$\text{large}$  & -$170.$ & -$359.$ \\ \hline
\end{tabular}
\end{center}
\vspace*{-1ex}
\end{table*}
\section{EXPERIMENTS}
\vspace*{-2ex}
\label{sec:exp}
Throughout this section, we consider spectral densities of the Mat\'ern class of covariance functions (order $1.5$ and length scale $1$), \citep[ch.~4]{Rasmussen2006}, and we consider eigenfunctions of the Laplace operator as the basis functions across all resolutions. Thus, for a $d_x$-dimensional input variable ${\vc x_{t}}$, we choose the basis functions, ${ \forall \vc x_t \in \vc x_{\TT_l\h j}}$,
\begin{multline*}
 {\!\!\!\! \phi_{i,l}\h j(\vc x_{t}, \vc \tau_l \h j)\!  = \!\prod_{d=1}^{d_x} ({1}/{\sqrt{\tau_{d,l}\h j}})\sin ({\pi i(x_{t,d} +\tau_{d,l}\h j)}/{2\tau_{d,l}\h j}) 
},\vspace*{-2ex}
 \end{multline*}
 with 
${ \lambda_{i}\h j(\vc \tau_l \h j) = \sum_{d=1}^{d_x}({\pi i}/{2\tau_{d,l}\h j})^2}$, ${ \forall i, l, j}$. 
The number of basis functions is set to ${p=\min\{n, 100\}}$.

In all experiments, we compare the performance of two different multiresolution model architectures, the conditionally independent and the fully independent models, namely $\mathrm{ciMRGP}$ and $\mathrm{fiMRGP}$. Note that $\mathrm{fiMRGP}$ here is obtained from $\mathrm{ciMRGP}$ by forcing the GPs across all resolutions to be independent (refer to Fig.~\ref{fig:model}).
For simplicity, we consider uniform subdivision of the index set by a factor of ${\mathfrak{q}=2}$. Finally, for instance, the notation $\mathrm{ciMRGP4}$ is used to refer to $\mathrm{ciMRGP}$ of resolution ${m=4}$.
\vspace*{-2ex}
\paragraph*{Conditional Independence versus Full Independence}
We begin with an illustrative experiment which demonstrate some limitations of the full independence assumption, non-smooth boundaries and overfitting in the sense of sensitivity to the chosen resolution. For this demonstration, we compare the performance of  $\mathrm{ciMRGP}$ and $\mathrm{fiMRGP}$ at various resolutions on synthetic data and real data.
Figure~\ref{fig:overfit}-a presents a regression task of identifying ($2$-dimensional) latent functions from $32$ noisy measurements on the $\mathrm{ToyData}$ dataset, App.~\ref{app:toy_data}. The dotted lines show the ground-truth and the solid lines indicate the predictions at $10^5$ test locations within the input range. At resolution ${m=1}$, the two models $\mathrm{ciMRGP1}$ and $\mathrm{fiMRGP1}$ perform comparatively. However, with increasing resolution, these models perform very differently. In particular, notice the non-smooth boundaries in the case of fully independent model at the highest resolution, 
$\mathrm{fiMRGP5}$, which are almost non-existing in $\mathrm{ciMRGP5}$. Given that the training set includes $n=32$ data samples, at ${m=5}$ practically every single data point is a region, $|\TT_l\h 5| =1, \forall l$. Also notice that $\mathrm{fiMRGP5}$ is closely following these data points, exhibiting signs of overfitting.
The overfitting issue associated with $\mathrm{fiMRGP}$ is partly due to the unconstrained flexibility of the GPs which manifest itself at the higher resolutions where the size of the regions under consideration becomes increasingly smaller. In our experiments on real data, however, the overfitting even happened at the lower resolutions. An example on the $\mathrm{vicon2}$ dataset, a subset of data recorded from a magnetic field, App.~\ref{app:vicon2}, is shown in Fig.~\ref{fig:overfit}-b. The $3$-dimensional noisy measurements are shown by dotted lines and the predicted strength of the magnetic fields at three different heights is estimated by each method and shown with solid lines. At ${m=1}$, both models ($\mathrm{ciMRGP1}$ and $\mathrm{fiMRGP1}$) perform equally well, but with the increase of resolution to $m=2$, $\mathrm{fiMRGP2}$ begins to fail which worsens as the resolution is further increased, while the $\mathrm{ciMRGP}$ family of models remain intact and comparative at all resolutions. 
\vspace*{-2ex}
\paragraph*{Regression on Multiple Datasets} We now compare the performance of various MRGP models on a number of datasets in a more structured fashion. As baselines, we include other scalable GP methods in this comparison. Key features of the datasets and models  are summarized in Table~1, and they are described in more details in App.~\ref{app:exp_data_method}. 
The performance is evaluated in terms of the root-mean-square error (RMSE) and the mean log-likelihood (MLL) on test sets, shown in Table~2 and Table~3, respectively. The model $\mathrm{ciMRGP8}$ is only applied to the datasets with larger data samples. The main results are summarized as follows. 
In the case of $\mathrm{ciMRGP}$, increasing the resolution from ${m=0}$ to the higher resolutions, ${m\geq 1}$, resulted in noticeable improvements in terms of MLL scores. The advantage is noticeable to a lesser degree in terms of the RMSE scores. In some cases, $\mathrm{fiMRGP}$ showed instabilities in particular at the higher resolutions ${m\geq2}$. In other cases, it only resulted in marginal improvements over the base model, $\mathrm{MRGP0}$. In comparison to the family of sparse GP models, $\mathrm{ciMRGP}$ at the higher resolutions performed well in terms of RMSE, but resulted in noticeably higher MLL scores. Generally, in cases with more data samples, we found it beneficial to increase the resolution to higher values. Consider the two datasets $\mathrm{scm1d}$ and $\mathrm{scm20d}$. We increased the resolution further to ${m=10}$. The resulting models $\mathrm{ciMRGP10}$ improved upon previously achieved scores reaching to RMSE and MLL scores of $0.994$ and $-6.4$ in the case of $\mathrm{scm1d}$, and $0.989$ and $-4.9$ in the case of $\mathrm{scm20d}$.
This additional gain of course comes with the cost of a longer computational time which may be justifiable in certain applications and for larger datasets. 


\vspace{-1ex}
\section{CONCLUSION}
\vspace*{-2ex}
We have derived a multiresolution Gaussian process model which assumes \emph{conditional independence} among the GPs across all resolutions. Relaxing the full independence assumption was shown to result in models robust to overfitting in the sense of reduced sensitivity to the chosen resolution, and predictions which are smooth at the boundaries.
Although models with high resolutions may safely be used for small amounts of data, they are most relevant, and computationally justified, when there are large amounts of data. This property, combined with the favorable computational advantages of the low rank representation via the Karhunen-Lo\`eve expansion, could make the proposed model appealing for large datasets.
 We conclude the paper by reiterating that sharing the basis axes is an effective approach toward creating cross-talk between GPs, an approach that could be useful for learning deep GPs with conditional independence across layers. 




\subsubsection*{Acknowledgements}
This research is financially supported by The Knut and Alice Wallenberg
Foundation (J. Taghia, contract number: KAW2014.0392), and by the Swedish Research Council (VR) via the project \emph{NewLEADS - New Directions in Learning Dynamical Systems} (T. Sch\"{o}n, contract number: 621-2016-06079). We are grateful for the help and equipment provided by the UAS Technologies Lab, Artificial Intelligence and Integrated Computer Systems Division (AIICS) at the Department of Computer and Information Science (IDA), Link\"{o}ping University, Sweden. The real data set used in this paper has been collected by Arno Solin, Niklas Wahlstr\"om, Manon Kok, and Simo S\"arkk\"a. We thank them for allowing us to use this data. We also thank Arne Leijon, Andreas Svensson, and Niklas Wahlstr\"om for useful feedback on early versions of this paper.
\bibliography{ref}
\bibliographystyle{abbrvnat}
\clearpage
\onecolumn
\appendix
\numberwithin{equation}{section}
\section{Prior model}
\label{app:prior}
This section describes our choice of the prior model parameters, and details of their initializations. 
\subsection{Prior over basis-axis scales} 
\label{app:prior_scale}
We assign a product of zero-mean Gaussian densities conditional on the basis-axis scale-precision variables as the prior over basis-axis scales,
\begin{equation}
\label{eq:pa}
p(\underline{\vc a} \mid {\vc r}) = \prod_{j=0}^m \prod_{l=1}^{|\TT^{(j)} |} \prod_{i=1}^p \mathcal{N}\left(a_{i,l}\h j; 0, {\left(\frac{r_i}{S\h j\Big(\sqrt{{\lambda_{i}\h j(\vc \tau_l\h j)}}\Big)}\right)}^{-1} \right), 
\end{equation}
where $S\h j \big(\cdot \big)$ is the spectral density of the covariance function and $\lambda_{i}\h j(\vc \tau_l \h j)$ is the eigenvalue of the basis function ${\phi_{i}\h j(\cdot)}$ at resolution $j$.
There are various choices of covariance functions \citep{Rasmussen2006}. Among them, we are interested in those for which ${S(\nu)\!\rightarrow\! 0}$ for all ${\nu\!\rightarrow \!\infty}$, that is the case for most classes of covariance functions, including Mat\'ern and exponentiated quadratic covariance functions. We have indicated spectral densities with indexing on $j$, as in general, we are free to choose different covariance functions at different resolutions. 
 Similarly, there are various choices of basis functions which are interpretable as GPs. As discussed in the paper, the choice of basis functions can in general be resolution-specific.


Our choice of prior implies that ${a_{i,l}^{(j)}}$ are resolution-region specific, which means that regardless of the resolution or the region the prior must be initialized with zero-mean even though the posterior mean is non-zero. 
\subsection{Prior over basis axes}
\label{app:prior_axis}
 Considering the possible index permutation across resolutions, we assign a  product of independent Bingham densities \citep{Bingham1974,Mardia2009}, conditional on the binary index-mapping matrix $\bc\Gamma$, as the prior over basis axes
\begin{gather}
\label{eq:pu}
\begin{split}
p(\vc U\mid \vc \Gamma) = \prod_{i=1}^p \prod_{k=1}^p \Big[
\mathcal{B}\big(\vc u_i; \ \bc B_k^{\prime} \big)
\Big]^{\Gamma_{ik}}=\prod_{i=1}^p \prod_{k=1}^p \left[ \frac{1}{\mathcal{C}(\vc \kappa_k^{\prime})} \exp\left( \vc u_i^\top \bc B_k^{\prime} \vc u_i \right)\right]^{\Gamma_{ik}}. 
\end{split}
\end{gather}
Here, 
 ${\bc B_k^{\prime}=\bc M_k^{\prime} \times \mathrm{diag}[\vc \kappa_k^{\prime}]\times {\bc M_k^{\prime}}^\top}$, and 
  the pair of ${\bc M_k^{\prime}=(\vc \mu_{k1}^{\prime}, \ldots, \vc \mu_{kd_y}^{\prime})}$, $ {\vc \mu_{kd_y}^{\prime}\in \mathcal{S}^{d_y-1}}$, ${\vc \kappa_k^{\prime}=(\kappa_{k1}^{\prime}, \ldots,\kappa_{kd_y}^{\prime})^{\top}}$ are given by the eigendecomposition of $\bc B_k^{\prime}$ and 
$\mathcal{C}(\vc \kappa_k^{\prime})$ is the Bingham normalization factor, which is algebraically problematic, but the saddle-point approximation \citep{Kume2005} provides an accurate numerical result. 

Notice that, at resolution ${j>0}$, $\bc B_k^{\prime}$ is given by the posterior hyper-parameter from the previous resolution~${j-1}$. At resolution $j=0$, we set simply ${\bc B_k^{\prime}= \bc B_0 = \vc 0}$. 
\subsection{Prior over basis-axis scale-precision} 
\label{app:prior_r}
Considering the possible index permutation across resolutions, we express the prior over precision of the basis scales as conditional on the binary index-mapping matrix $\vc \Gamma$ using Gamma densities
\begin{equation}
\label{eq:prior_lambda}
p(\vc r \mid \vc \Gamma) = \prod_{i=1}^p\prod_{k=1}^p \left[\mathcal{G}\big(r_i; \ \alpha_k^{\prime}, \beta_k^{\prime} \big)\right]^{\Gamma_{ik}}, 
\end{equation}
where $\alpha_k^{\prime}$ and $\beta_k^{\prime}$ are the Gamma densities shape and inverse scale hyper-parameters. At resolution ${j>0}$, $\alpha_k^{\prime}$ and $\beta_k^{\prime}$ are the posterior hyper-parameters computed from resolution $j-1$. At ${j=0}$, in absence of prior data, non-informative distributions may be assigned with ${\alpha_k^{\prime}\rightarrow 0}$, but ${\beta_k^{\prime}}$ may still be assigned an informative value.  
Values of~${\beta_k^{\prime}}$ for which ${\alpha_k^{\prime}/ \beta_k^{\prime} \rightarrow 0}$ reduces the overall influence of the prior toward a non-regularized basis function expansion. 
\subsection{Prior over basis-axis index mapping}
\label{app:prior_map}
As discussed earlier, the index-mapping binary matrix $\vc \Gamma$ has exactly one element $\Gamma_{ik}=1$ in each row and each column, indicating that the basis axis identified by index $k$ in the previous resolution ${j-1}$ is identical to the basis axis denoted by index~$i$ at the current resolution~$j$. The prior probability mass for these index-mapping variables is assigned as totally non-informative, except for the uniqueness requirement 
\begin{subequations}
\label{eq:priorZ}
\begin{align} 
\sum_{k =1}^p p(\Gamma_{ik}=1) &= 1, \qquad \forall i \in \{1,\ldots, p\},\\ \vspace{-4ex}
\sum_{i =1}^p p(\Gamma_{ik}=1) &= 1, \qquad \forall k \in \{1,\ldots, p\}.
\end{align}
\end{subequations}
\subsection{Prior over overall bias and residual noise precision}
\label{app:prior_bias}
We assign product of Gaussian-Gamma densities over the joint distribution of the overall bias and the residual noise precision as
\begin{align}
p(\underline{\vc b} , \underline{\gamma}) = \prod_{j=0}^m \prod_{l=1}^{|\TT^{(j)} |} \mathcal{N}\Big(\vc b_l\h j;  {\vc \nu_{\mathfrak 0}}_l\h j, \frac{1}{{\vartheta_{\mathfrak 0}}_l \h j \gamma_l\h j} \Big)  \mathcal{G}\Big(\!\gamma_l\h j; \mathfrak{c_0}_l\h j, \mathfrak{d_0}_l \h j \Big).
\end{align}
In the absence of prior information, a non-informative prior must be applied by setting ${{\vc \nu_{\mathfrak 0}}_l\h j=\vc 0}$ and ${\vartheta_{\mathfrak 0}}_l \h j\!\rightarrow \! 0$. The hyper-parameters $\mathfrak{c_0}_l\h j$ and $\mathfrak{d_0}_l\h j$ are shape and inverse scale parameters of the corresponding Gamma distributions. In the absence of prior information, a noninformative distribution is assigned by ${\mathfrak{c_0}_l\h j\rightarrow 0}$ , but ${\mathfrak{d_0}_l\h j}$ may still be assigned an informative value to indicate the most likely value (mode), ${{\mathfrak{d_0}_l\h j}/{(\mathfrak{c_0}_l\h j +1)}}$, for the residual variance which has an inverse-gamma distribution.
\section{Posterior model}
\label{app:post}
In this section, we summarize the optimized posterior distribution which is obtained by maximizing the lower bound~$\mathcal{L}$ in~\eqref{eq:LL}. For ease of notation, we use: ${\la \ \cdot \ \ra_{q(\cdot)}\equiv \la \ \cdot \ \ra}$ wherever possible. Descriptive statistics of the posterior distributions are summarized in Appendix~\ref{app:stats}. 
\subsection{Conditional posterior over basis-axis scales}
\label{app:post_scale}
Optimized conditional posterior distribution of $q(\underline{\vc a} \!\mid\! \vc U) $ is given by the following product of Gaussian densities
\begin{align*}
q(\underline{\vc a} \!\mid\! \vc U) = \prod_{j=0}^m \prod_{l=1}^{|\TT^{(j)} |} \prod_{i =1}^p \mathcal{N}\big(a_{i,l}\h j; \  \mathfrak{m}_{i,l}\h j(\vc u_i), {\mathfrak{v}_{i,l}\h j}^{-1} \big),  
\end{align*}
with the mean value $\mathfrak m_{i,l}^{(j)}(\vc u_i)$ as the function of the basis axis vector $\vc u_i$ and the precision $\mathfrak v_{i,l}^{(j)}$ given by
\begin{align*}
&\mathfrak v_{i,l}^{(j)} =\frac{\la r_i\h j\ra}{S\h j\Big(\sqrt{{\lambda_{i}\h j(\widehat{\vc \tau}_l\h j)}}\Big)} + \la \gamma_{l}\res\ra \!\!\sum_{t \in \TT_l^{(j)}} \left({\phi_i\h j(\vc x_t, \widehat{\vc \tau}_l \h j)}\right)^2 , \\
&\mathfrak m_{i,l}^{(j)}(\vc u_i) = \zeta^{(j)}_{i,l} \vc u_i^\top \widetilde{\vc z}^{(j)}_{i,l},
\\
\end{align*}
where we have defined
\begin{align*}
&{\zeta_{i,l}^{(j)} = \frac{\la\gamma_l\res\ra}{\mathfrak v_{i,l}^{(j)}}},
\\
&\widetilde{\vc z}^{(j)}_{i,l} \!=\!\left\{
\begin{array}{ll}
 \widetilde{\vc y}_{i}= \sum_{t \in \TT^{(0)}}\phi_{i}\h 0 (\vc x_t, \widehat{\vc \tau}\h 0)\Big(\vc y_{t} - \bar{\widetilde{\vc y}}_{i}\Big), & \forall j=0\\
 \sum_{t \in \TT_l^{(j)}}\phi_{i}\h j (\vc x_t, \widehat{\vc \tau}_l \h j)\Big(\!\!\la \vc z_{t,l}^{(j)}\ra - \bar{\widetilde{\vc z}}^{(j)}_{i,t,l}\Big), & \forall j\geq 1
\end{array}
              \right.,
\\
&{\bar{\widetilde{\vc y}}_{i} = \la\vc b \ra +\sum_{k\neq i}^p \la a_{k}\vc u_k\ra \phi_k\h 0(\vc x_t, \widehat{\vc \tau}\h 0)}, \\
&{\bar{\widetilde{\vc z}}^{(j)}_{i,t,l} = \Bigg(\sum_{j^{\prime}=0}^{j-1} \bar{\bc f}_{t,l}^{j^{\prime}} \Bigg)  + \la \vc b_l\h j \ra + \sum_{k\neq i}^p \la a_{k,l}^{(j)} \vc u_k\ra \phi_k\h j(\vc x_t,\widehat{\vc \tau}_l \h j)}.
\end{align*}
The seemingly complicated form of this result makes intuitively good sense: The conditional expected value of the basis-axis scale variables, given by $\mathfrak m_{i,l}^{(j)}(\vc u_i)$, is determined by the mean predictions from the previous resolution plus the remaining part of the observed (or latent for ${j>0}$) vector that is not already explained by its components along the other basis axes. The conditional expected value is scaled by $\zeta_{i,l}^{(j)}$, which is the currently estimated proportion of the variance of the observed data (or latent variables for ${j>0}$) that is explained by the basis-axis scale variables in the $i$th axis, in relation to the total variance that also includes the residual noise component along this axis.
\subsection{Posterior over basis axes}
\label{app:post_axis}
Given a posterior distribution ${q(\vc a_l^{(j)} \mid \vc U)}$ and using $q(\vc a_l^{(j)}, \vc U) = q(\vc U)q(\vc a_l^{(j)} \mid \vc U)$, it can be shown that the optimized posterior distribution ${q(\vc U)}$ is given by the product of Bingham densities 
\begin{align*}
\begin{split}
q(\vc U) = 
\prod_{i=1}^p\mathcal{B}\big(\vc u_i;\ \bc B_i \big)
=\prod_{i=1}^p \frac{1}{\mathcal{C}(\vc \kappa_i)} \exp\left( \vc u_i^\top \bc M_i \times \mathrm{diag}[\vc \kappa_i] \times \bc M_i^\top \vc u_i \right), 
\end{split}
\end{align*}
where as before the pair of $\vc \kappa_i$ and $\bc M_i$ are eigenvalues and the corresponding eigenvectors of
\begin{align*}
\label{eq:B_i}
\bc B_i  =  \left(\sum_{k=1}^p \! \la \Gamma_{ik} \ra \bc B^{\prime}_k\right) +  \sum_{l=1}^{|\TT^{(j)} |}  \frac{\la\gamma_{l}\res\ra}{2}\zeta_{i,l}^{(j)}  \widetilde{\vc z}^{(j)}_{i,l} {\widetilde{\vc z}^{(j)}_{i,l}\! ~}^{\top}.
\end{align*}
Note the first term where Bingham's posterior hyper-parameter from the previous resolution, $\bc B_k^{\prime}$, has been weighted by ${ \la \Gamma_{ik} \ra}$. This ensures that the axis indices remain aligned throughout and hence allows for recursive (successive) learning of these parameters.
\subsection{Posterior over basis-scale precision} 
\label{app:post_r}
The optimized posterior distribution of the latent variables $\vc r$ is given by the product of Gamma densities as
\begin{align*}
{q({\vc r}) = \prod_{i=1}^p \mathcal{G}\big ( r_i; \ {\alpha_i} , {\beta_i}\big )},
\end{align*}
where $\alpha_i$ and $\beta_i$ are the shape and inverse scale posterior hyper-parameters of Gamma density given by
\begin{align*}
&\alpha_i = \left(\sum_{k=1}^p \la \Gamma_{ik} \ra \alpha_{k}^{\prime} \right) + \frac{|\TT^{(j)}|}{2}, \\
& \beta_i = \left( \sum_{k=1}^p \la \Gamma_{ik} \ra \beta_{k}^{\prime}  \right) + \frac{1}{2}\sum_{l=1}^{|\TT^{(j)}|} \frac{\la \big(a_{i,l}^{(j)}\big)^2 \ra}{S\h j\Big(\sqrt{{\lambda_{i}\h j(\widehat{\vc \tau}_l\h j)}}\Big)},
\end{align*}
where $\alpha_{k}^{\prime}$ and $\beta_{k}^{\prime}$ are posterior hyper-parameters from the previous resolution, $k$, weighted by the posterior mean of the basis-axis index-mapping variable, ${ \la \Gamma_{ik} \ra}$.
\subsection{Posterior over basis-axis index mapping} 
\label{app:post_map}
The optimized posterior distribution of the latent variables $q(\vc \Gamma)$ is given by 
${q(\vc \Gamma) = \prod_{i=1}^p \prod_{k=1}^{p} \omega_{ik}^{\Gamma_{ik}}}$,
where the probability parameters are normalized using scale factors $\eta_i$ and $\eta_k$ as
\begin{align*}
&\omega_{ik} = \eta_i \eta_k \widetilde{\omega}_{ik}, \\   &\mathrm{such~that:} 
\left\{ 
                \begin{array}{ll}
                  \sum_{k=1}^p \omega_{ik}=1, \ \forall i\in \{1,\ldots,p\}\\
                  \sum_{i=1}^p \omega_{ik}=1, \ \forall k\in \{1,\ldots,p\}
                \end{array}
              \right., \notag
\end{align*}
to satisfy the prior requirements, Eq.~\eqref{eq:priorZ}, with 
\begin{align*}
\log \widetilde{\omega}_{ik} = \la \vc u_i^\top \bc B_k^{\prime} \vc u_i \ra - \log \mathcal{C}(\bc B_k^{\prime}) 
+ \alpha_k^{\prime} \log \beta_k^{\prime} - \log \digamma(\alpha_k^{\prime}) + (\alpha_k^{\prime} - 1)\la \log r_i\ra - \beta_k^{\prime}\la r_i \ra,
\end{align*}
where $\digamma(\cdot)$ denotes the digamma function. 
We may view $\log \widetilde{\omega}_{ik}$ as a logarithmic similarity measure between the $k$th prior axes at the previous resolution and $i$th posterior axes at the current resolution.
\subsection{Posterior distribution of the latent remainder term}
\label{app:post_remain}
The optimal posterior distribution of $q(\underline{\zt})$ is given by
\begin{align*}
&q(\underline{\zt}) =\prod_{j=1}^m \!\prod_{l=1}^{|\TT\h j|}  \prod_{t \in \TT_l\h j}\mathcal{N}\Big(\vc z_{t,l}\h j;  \la \bar{\vc z}_{t,l}\h j\ra, \la{\!\gamma\h j\!}\ra^{\!-1} \Big), \\
&\la \bar{\vc z}_{t,l}\h j\ra=\sum_{j^{\prime}=0}^{j-1}  \bar{\bc f}_{t,l}^{j^{\prime}}  + \la \vc b_l\h j \ra + \sum_{i=1}^p \la a_{i,l}\h j \vc u_i\ra \phi_{i}\h j(\vc x_t, \widehat{\vc \tau}_l \h j).
\end{align*}
\subsection{Posterior distribution of overall bias and residual noise precision}
\label{app:post_bias}
The optimized posterior of the joint distribution of the mean vector and the residual noise is given by
\begin{align*}
q(\underline{\vc b} , \underline{\gamma}) = \prod_{j=0}^m \prod_{l=1}^{|\TT^{(j)} |} \mathcal{N}\Big( \vc b_l\h j;  \vc \nu_l\h j, \frac{1}{\vartheta_l \h j \gamma_l\h j} \Big) \mathcal{G}\Big(\gamma_l\h j; \mathfrak{c}_l\h j, \mathfrak{d}_l \h j  \Big),
\end{align*}
with the posterior hyper-parameters given by
\begin{align*}
&\vartheta_l\h j = {\vartheta_{\mathfrak 0}}_l\h j + |\TT_l\h j|, \\
\\
&\vc \nu_l\h j =\frac{1}{{\vartheta}_l\h j}\left({\vartheta_{\mathfrak 0}}_l\h j\vc {\nu_{\mathfrak 0}}_l\h j + \bar{\vc \nu}_l\h j\right), \\
&\mathfrak{c}_l\h j = \mathfrak{c_0}_l\h j + \frac{d_y}{2} |\TT_l\h j|, \\
&\mathfrak{d}_l\h j = \mathfrak{d_0}_l\h j + \frac{1}{2}\bar{\mathfrak{d}}_l\h j,
\end{align*}
where $\bar{\vc \nu}\h 0$ and $\bar{\mathfrak{d}}\h 0$ are given by
\begin{align*}
&\bar{\vc \nu}\h 0  = \sum_{t\in \TT\h 0} \left( \vc y_{t}  - \sum_{i=1}^p \la a_{i} \vc u_i \ra \phi_i\h 0(\vc x_t, \widehat{\vc \tau}\h 0) \right),
\\
&\bar{\mathfrak{d}}\h 0 = {\vartheta_{\mathfrak 0}}\h 0 \| \vc {\nu_{\mathfrak 0}}\h 0  \big \|^2 - {\vartheta\h 0} \| \vc \nu\h 0 \big\|^2 +  \\ 
& \quad + \sum_{t\in \TT\h 0}\! \!\Bigg(\!\Big\|  \vc y_{t} - \sum_{i=1}^p \phi_i\h 0(\vc x_t, \vc \tau\h 0) \la a_{i} \vc u_i\ra  \Big \|^2  
+ \sum_{i=1}^p \Big({\phi_i\h 0(\vc x_t, \vc \tau\h 0)}\Big)^2 \la\big\| a_{i} \vc u_i- \la a_{i} \vc u_i\ra \big\|^2 \ra \Bigg).
\end{align*}
and similarly $\bar{\vc \nu}_l\h j$ and $\bar{\mathfrak{d}}_l\h j$, $\forall j\geq 1$, are given by
\begin{align*}
&\bar{\vc \nu}_l\h j  = \sum_{t\in \TT_l\h j}\Bigg( \la \vc z_{t,l} \h j \ra - \sum_{j^{\prime}=0}^{j-1} \bar{\bc f}_{t,l}\h {j^{\prime}} - \sum_{i=1}^p \la a_{i,l}\h j \vc u_i \ra \phi_i\h j(\vc x_t, \widehat{\vc \tau}_l\h j) \Bigg),
\\
&\bar{\mathfrak{d}}_l\h j = {\vartheta_{\mathfrak 0}}_l\h j \| \vc {\nu_{\mathfrak 0}}_l\h j  \big \|^2 - {\vartheta_l\h j} \| \vc \nu_l\h j \big\|^2 +  \\ 
& + \sum_{t\in \TT_l\h j}\! \!\Bigg(\!\Big\| \la \vc z_{t,l}\h j\ra - \sum_{i=1}^p \phi_i\h j(\vc x_t, \vc \tau_l\h j) \la a_{i,l}\h j \vc u_i\ra  -  \sum_{j^{\prime}=0}^{j-1} \bar{\bc f}_{t,l}\h {j^{\prime}}\Big \|^2  \\ &\quad + \la \left \| \vc z_{t,l}\h j - \la \vc z_{t,l}\h j\ra\right \|^2 \ra+  \!\sum_{j^{\prime}=0}^{j-1}  \mathrm{E}\Big [ \left \|
 {\bc f}_{t,l}\h {j^{\prime}} -  \bar{\bc f}_{t,l}\h {j^{\prime}} \right \|^2 \Big] \\
&\quad \quad
+ \sum_{i=1}^p \Big({\phi_i\h j(\vc x_t, \vc \tau_l\h j)}\Big)^2 \la\big\| a_{i,l}\h j \vc u_i- \la a_{i,l}\h j \vc u_i\ra \big\|^2 \ra\!\Bigg).
\end{align*}
Estimating the noise precision at resolution~${j\geq1}$ also includes the second central moments of the predictive processes and the latent remainder terms at the previous resolutions.
\section{Descriptive statistics}
\label{app:stats}
Descriptive statistics of the posterior distributions $q(r_i)$, ${q( a_{i,l}^{(j)}\given \vc u_i)}$, and $q(\ztj j)$ are conveniently given by the known statistics of the Gamma and Gaussian distributions. For $q(\vc \Gamma)$, we have the standard result of  ${ \la \Gamma_{ik} \ra = \omega_{ik}}$. With a special notational treatment for $j=0$, the required statistics for the joint posterior 
 $q(\vc u_i, a_{i,l}^{(j)})$, $\forall j$, are summarized as 
\label{eq:stats}
\begin{align*}
  &\la \vc u_i \vc u_i^{\top} \ra = \sum_{d=1}^{d_{y}} \rho_{id}(\vc \kappa_{i}) \vc \mu_{id} \vc \mu_{id}^{\top},
  \\
  &\la  a_{i,l}^{(j)} \vc u_i \ra = \zeta_{i,l}^{(j)} \la \vc u_i \vc u_i^{\top} \ra \widetilde{\vc z}^{(j)}_{i,l},
   \\
   &\la \left(a_{i,l}^{(j)}\right)^2\ra=  \frac{1}{\mathfrak v_{i,l}^{(j)}} + \left({\zeta_{i,l}\h j}\right)^2 
   {\widetilde{\vc z}^{(j)}_{i,l}\ }^{\top} \la \vc u_i \vc u_i^{\top} \ra {\widetilde{\vc z}^{(j)}_{i,l}},
   \\&
   \la \left \| a_{i,l}\h j \vc u_i - \la  a_{i,l}\h j\vc u_i\ra \right \|^2 \ra  =  \frac{1}{\mathfrak v_{i,l}^{(j)}} + {\zeta_{i,l}\h j}^2 \left({\widetilde{\vc z}^{(j)}_{i,l}}\right)^{\top} \Big( \la \vc u_i \vc u_i^{\top} \ra - \la \vc u_i \vc u_i^{\top} \ra \la \vc u_i \vc u_i^{\top} \ra \Big){\widetilde{\vc z}^{(j)}_{i,l}},
\end{align*}
where $\rho_{id}(\vc \kappa_i)$ is the $d$-th element of $\vc \rho_{i}(\vc \kappa_i)$ given by
\begin{align*}
\vc \rho_{i}(\vc \kappa_i) = \frac{\partial \log \mathcal{C}(\vc \kappa_i)}{\partial \vc \kappa_i}, \quad \forall i\in \mathcal{P}.
\end{align*}
The saddle-point approximation of \citet{Kume2005} is used to calculate the derivatives above.

The mean and the second central moment of the predictive processes can be computed using Eq.~\eqref{eq:pred_m} and \eqref{eq:pred_c},
\begin{align*}
&\mathrm{E}\left[ {\bc f}_{t,l}\h j \right] \equiv \bar{\bc f}_{t,l}\h j= {\bar{\bc f}_l\h j(\vc x_t)}, \\
& 
\mathrm{E}\left[ \big \| {\bc f}_{t,l}\h j - \bar{\bc f}_{t,l}\h j \big\|^2\right] =\mathrm{trace}\left[ {\bar{\bc F}_l\h j}(\vc x_t)\right], \quad \forall \vc x_t\in \TT_l\h j.
\end{align*}
\section{Predictive process}
\label{app:pred}
For a new test input $\vc x^*$, we shall first determine if we know to which region it belongs in each resolution. If such information is available the required statistics of the approximate predictive process at $\vc x^*$ can be computed from the sum of their contributions across all resolutions, as
\begin{subequations}
\label{eq:pred}
\begin{align}
&\mathrm{E}\left[p(\vc f(\vc x^*) \mid  \vc x^*, \underline{\vc x_{\TT}}, \yt, \underline{\vc z_{\TT}}) \right] =  \sum_{j=0}^m {\bar{\bc f}\h j(\vc x^*)},
\\
&\mathrm{Cov}\left[ p(\vc f(\vc x^*)\mid\vc x^*, \underline{\vc x_{\TT}}, \yt,\underline{\vc z_{\TT}}) \right] = \sum_{j=0}^m {\bar{\bc F}\h j}(\vc x^*),
\end{align}
where ${\bar{\bc f}_l\h j(\vc x^*)}$ and ${\bar{\bc F}_l\h j}(\vc x^*)$ are given by
\begin{align}
\label{eq:pred_m}
{\bar{\bc f}_l\h j(\vc x^*)} = \la\vc b_l\h j\ra +\! \sum_{i=1}^p \la a_{i,l}^{(j)} \vc u_i \ra \phi_i\h j(\vc x^*, \widehat{\vc \tau}_l \h j), 
\end{align}
\vspace{-2ex}
\begin{multline}
\label{eq:pred_c}
{\bar{\bc F}_l\h j}(\vc x^*) = \la {\vc b_l\h j}{\vc b_l\h j}^{\top} \ra - \la {\vc b_l\h j}\ra \la {\vc b_l\h j}\ra^{\top} + 
\\ + \sum_{i=1}^p 
{\left( \phi_i\h j (\vc x^*,  \widehat{\vc \tau}_l \h j) \right)}^{ 2} \times  \Bigg[
\la \left(a_{i,l}^{(j)}\right)^{2}\ra \la \vc u_i \vc u_i^{\top} \ra   
-\la a_{i,l}^{(j)} \vc u_i \ra \la  a_{i,l}^{(j)} \vc u_i \ra^{\top} \Bigg]. 
\end{multline}
\end{subequations}
In many applications however we may indeed not know the position of $\vc x^*$ in the training index sets, $\TT\h j, \forall j,$---in other words we may not know to which region $\vc x^*$ belongs at a given resolution. In such cases, since the basis axes are shared across all resolutions and learnt in a group fashion, predictions are made only from ${j=0}$,
\begin{subequations}
\label{eq:pred_0}
\begin{align}
&\mathrm{E}\left[p(\vc f(\vc x^*) \mid  \vc x^*, \underline{\vc x_{\TT}}, \yt, \underline{\vc z_{\TT}}) \right] =   {\bar{\bc f}\h 0(\vc x^*)},
\\
&\mathrm{Cov}\left[ p(\vc f(\vc x^*)\mid\vc x^*, \underline{\vc x_{\TT}}, \yt, \underline{\vc z_{\TT}}) \right] =  {\bar{\bc F}\h 0}(\vc x^*).
\end{align}
\end{subequations}
We emphasize that, among others, this is one of the advantages of the conditional independence over models with full independence. 
\section{Optimization of basis interval variables}
\label{app:tau}
The basis interval variables ${\vc \tau_l \h j=(\tau_{1,l}\h j, \ldots, \tau_{d_x, l}\h j)^{\top}}$ are optimized using maximum likelihood estimation, as an analytical solution within our standard variational inference may not exist in general form for various choices of basis functions and spectral densities. The optimized point estimate values are given from
\begin{align*}
\widehat{\tau}_{d,l} \h j &= \argmax_{{\tau}_{d,l} \h j}  \  h(\tau_{d,l}\h j),  \\ \mathrm{s.t. \quad } &{L_{d_x, l}\h j\!<\!{\tau}_{d,l} \h j\! <\!L_{d_x, l}\h j + \frac{p}{L_{d_x, l}\h j}},
\end{align*}
where $L_{d_x, l}\h j$ is the input range at $(j,l)$, and
\begin{align*}
h(\tau_{d,l}\h j) \propto h_{\mathrm{prior}}(\tau_{d,l}\h j) + h_{\mathrm{likelihood}}(\tau_{d,l}\h j ),
\end{align*}
where $h_{\mathrm{prior}}(\tau_{d,l}\h j)$ includes all relevant terms from the prior, 
\begin{align*}
h_{\mathrm{prior}}(\tau_{d,l}\h j)  = -\frac{1}{2}\sum_{i=1}^p \Bigg(\log \left(\mathfrak{S}_{d, l, i}\h j(\tau_{d,l}\h j)  \right)  - \frac{\la r_i \ra \la \left(a_{i,l}^{(j)}\right)^2\ra}{2 \mathfrak{S}_{d, l, i}\h j(\tau_{d,l}\h j) } \Bigg) ,
\end{align*}
and $h_{\mathrm{likelihood}}(\tau_{d,l}\h j) $ includes all relevant terms in the likelihood term,
\begin{multline*}
h_{\mathrm{likelihood}}(\tau_{d,l}\h j)  = - \la\gamma_l\h j\ra\sum_{t=1}^{\TT_l \h j} \sum_{i=1}^p\Bigg( \Big\|\mathfrak{X}(x_{t,d, i}\h j, \tau_{d,l}\h j)\Big \|^{2} +2\left(\mathfrak{k}_{t,l}\h j\right)^{\top}  \mathfrak{X}(x_{t,d, i}\h j, \tau_{d,l}\h j)  \\+ \la \left \| a_{i,l}\h j \vc u_i - \la  a_{i,l}\h j\vc u_i\ra \right \|^2 \ra \Big(\tilde{ \phi}_{i,d,l,t} \h j \phi_i\h j(x_{t,d}, \tau_{d,l}\h j)\Big)^2 \Bigg),
\end{multline*}
where we have defined
\begin{align*}
& \mathfrak{k}_{t,l}\h j = \la b_l\h j \ra + \sum_{j^{\prime}=0}^{j-1}{\bar{\bc f}_{t,l}\h {j^{\prime}}} -\frac{1}{2}{\vc z_{t,l}\h j},
\\
& \mathfrak{S}_{d, l, i}\h j(\tau_{d,l}\h j) = S\h j\left( \sqrt{\lambda_i\h j(\tau_{d, l}\h j) + \tilde{\lambda}_{i, d, l}\h j}\right),
\\
&\mathfrak{X}(x_{t,d, i}\h j, \tau_{d,l}\h j) = \la a_{i,l}^{(j)} \vc u_i\ra \tilde{ \phi}_{i,d,l,t} \h j \phi_i\h j(x_{t,d}, \tau_{d,l}\h j),
\\
&\tilde{\lambda}_{i, d, l}\h j = \sum_{k\neq d}^{d_x} \lambda_i \h j(\widehat{\tau^{\prime}}_{k,l}\h j), \\
&\tilde{ \phi}_{i,d,l,t} \h j = \prod_{k\neq d}^{d_x} \phi_i\h j (x_{t,k} , \widehat{\tau^{\prime}}_{k, l}\h j),
 \end{align*}
 where $\widehat{\tau^{\prime}}_{k, l}\h j$ are the previous optimized values.
The optimization problem is solved numerically. 


\section{Algorithm}
\label{app:alg}
\begin{itemize}
\item Initialize the basis intervals $\bc \tau_l\h j$.
\item[1.] {Assign priors}
\begin{itemize}
\item Initialize the resolution-region-specific prior distributions ${p(\underline{\vc a}\given \vc r),  p(\underline{\vc b}, \underline{\gamma})}$ by setting their hyperparameters to the default values according to Appendices~\ref{app:prior_scale},~\ref{app:prior_bias}. During recursive learning the prior hyperparamters remain unaltered and will not be updated.
\item Initialize shared prior distributions ${p(\vc U\given \vc \Gamma), p(\vc r\given \vc \Gamma), p(\vc \Gamma)}$ according to Appendices~\ref{app:prior_axis},~\ref{app:prior_r},~\ref{app:prior_map}. During recursive learning at resolution ${j\!>\!1}$, the posterior hyperparameters from the previous resolution ${j\!-\!1}$ are used as the prior hyperparameters for the current resolution $j$. 
\end{itemize}
\item[2.] {Update posteriors}
\begin{itemize}
  \item Resolution-region-specific posteriors ${q(\underline{\vc a}\given \vc U),  q(\underline{\vc b}, \underline{\gamma})}$ are updated according to Appendices~\ref{app:post_scale},~\ref{app:post_bias}.
  \item Shared posteriors ${q(\vc U), p(\vc r), p(\vc \Gamma)}$ are updated according to Appendices~\ref{app:post_axis},~\ref{app:post_r},~\ref{app:post_map}.
\end{itemize}
\item[3.] If necessary, update the basis intervals according to Appendix~\ref{app:tau}. 
\item Repeat steps 1 to 3 until convergence criteria are met.
\end{itemize}

\section{Experiment details}
\label{app:exp_data_method}
This section provides further details on the experiments in Sec.~\ref{sec:exp}. 
\subsection{Datasets}
\label{app:data_exp}
\subsubsection{$\mathrm{oes10}$ and $\mathrm{oes97}$}
\label{app:oes}
The datasets $\mathrm{oes10}$ and $\mathrm{oes97}$ were obtained from \citep{Spyromitros-Xioufis2016}. The Occupational Employment Survey (OES) datasets contain records from the years of 1997 (OES97) and 2010 (OES10) of the annual Occupational Employment Survey compiled by the US Bureau of Labor Statistics. As described in \citep{Spyromitros-Xioufis2016}, "each row provides the estimated number of full-time equivalent employees across many employment types for a specific metropolitan area". We selected the same $16$ target variables as listed in \citep[Table~5]{Spyromitros-Xioufis2016}. The remaining $298$ and $263$ variables serve as the inputs in the case of $\mathrm{oes10}$ and $\mathrm{oes97}$, respectively. Data samples were randomly divided into training and test sets (refer to Table~1).
\subsubsection{$\mathrm{atp1d}$ and $\mathrm{atp7d}$}
\label{app:atp7d}
The datasets $\mathrm{atp1d}$ and $\mathrm{atp7d}$ were obtained from \citep{Spyromitros-Xioufis2016}. The Airline Ticket Price (ATP) dataset includes the prediction of airline ticket prices. As described in \citep{Spyromitros-Xioufis2016}, the target variables  are either the next day price, $\mathrm{atp1d}$, or minimum price observed over the next seven days $\mathrm{atp7d}$ for $6$ target flight preferences listed in \citep[Table~5]{Spyromitros-Xioufis2016}. There are $411$ input variables in each case. The inputs for each sample are values considered to be useful for prediction of the airline ticket prices for a specific departure date, for example, the number of days between the observation date and the departure date, or the boolean variables for day-of-the-week of the observation date. Data samples were randomly divided into training and test sets (refer to Table~1).
\subsubsection{$\mathrm{scm1d}$, $\mathrm{scm1d}\text{\normalfont -a}$ and $\mathrm{scm20d}$}
\label{app:scm}
The datasets $\mathrm{scm1d}$ and $\mathrm{scm20d}$ were obtained from \citep{Spyromitros-Xioufis2016}. The Supply Chain Management (SCM) datasets are derived from the Trading Agent Competition in Supply Chain Management (TAC SCM) tournament from 2010. As described in \citep{Spyromitros-Xioufis2016}, each row corresponds to an observation day in the tournament. There are $280$ input variables in these datasets which are observed prices for a specific tournament day. The datasets contain $16$ regression targets, where each target corresponds to the next day mean price  $\mathrm{scm1d}$ or mean price for 20 days in the future $\mathrm{scm20d}$ for each product \citep[Table~5]{Spyromitros-Xioufis2016}. Dataset $\mathrm{scm1d}$-a is a subset of $\mathrm{scm1d}$ which includes the first 3000 samples.
Data samples were randomly divided into training and test sets (refer to Table~1). 

\subsubsection{$\mathrm{naval}$}
\label{app:naval}
The dataset $\mathrm{naval}$ \citep{Coraddu2013} was obtained from UCI Machine Learning Repository\footnote{http://archive.ics.uci.edu/ml/datasets/condition+based+maintenance+of+naval+propulsion+plants}. The input variables are $16$-dimensional feature vectors containing the gas turbine (GT) measures at steady state of the physical asset, for example, GT rate of revolutions, and Gas Generator rate of revolutions. The targets are two dimensional vectors measuring GT Compressor decay state coefficients and GT Turbine decay state coefficients. Data samples were randomly divided into training and test sets (refer to Table~1).

\subsubsection{$\mathrm{vicon}$}
\label{app:vicon}
The dataset $\mathrm{vicon}$ contains measurements recorded from a magnetic field which maps a 3-dimensional (3D)  position to a 3D magnetic field strength \citep{Jidling2017}\footnote{More information about data can be found in \citep{Jidling2017}. The data is available from  https://github.com/carji475/linearly-constrained-gaussian-processes}. The inputs are $(x,y,z)$-coordinates and the responses measured at there different heights are the target values. Data samples were randomly divided into training and test sets (refer to Table~1). 

\subsubsection{$\mathrm{hrtf}$}
\label{app:hrtf}
The dataset $\mathrm{hrtf}$ was obtained from the CIPIC HRTF database \citep{algazi01} which is a public-domain database of high-spatial-resolution head-related transfer function (HRTF) measurements\footnote{Details of the database can be found at: https://www.ece.ucdavis.edu/cipic/spatial-sound/hrtf-data/.}. We used the datasets of $37$ subjects divided into training and test sets (refer to Table~1). Data for each subject includes $200$-dimensional measurements of head-related impulse responses (HRIRs) and $8$ input variables which are in fact the anthropometric parameters considered to have strong direct physical effect
on HRIRs. The objective is to predict the HRIRs of the test subjects given their individualized anthropometric parameters\footnote{The preprocessed data can be obtained from our GitHub page: <GitHub link to data>.}.    
\subsubsection{$\mathrm{nengo}$}
\label{app:nengo}
The dataset $\mathrm{nengo}$ for this analysis was generated using The neural engineering object (Nengo) simulator \citep{Trevor12014,Taghia2018}. The generated time series data is constructed from a Nengo-based spiking model of action selection in the cortex-basal ganglia-thalamus circuit with timing predictions that are well matched to both single-cell recordings in rats and psychological paradigms in humans. Target measurements here are ensembles of leaky integrate-and-fire neurons comprised from seven nodes of the basal ganglia circuit (namely: globus pallidus internal, globus pallidus external, subthalamic nucleus, striatum D1, striatum D2; thalamus; motor cortex). Measurements from these $7$ nodes are the target outputs. The advantage of using the Nengo neural simulator in the regression task is that we also have access to the ground-truth, the function generating the noisy target measurements at each node. Data samples\footnote{Data can be downloaded from our GitHib page: <GitHub link to data>.} were randomly divided into training and test sets (refer to Table~1).
\subsubsection{\texorpdfstring{$\mathrm{lorenz96}$}{Lg}}
\label{app:lorenz}
The synthetic dataset $\mathrm{lorenz96}$ was generated using the Lorenz model \citep[Eq. 3.2]{Lorenz96}. Using a locally defined notation, consider the Lorenz model of
$${\frac{\partial x_k}{\partial t} = -x_{k-1}(x_{k - 2} - x_{k+1})  - x_{k} + F}, \qquad {\forall k \in K},$$ where $x_k$ represent the state of the system and $F$ is the forcing constant. 
In our simulation, 
we let ${K=20}$ and set ${F=8}$, which cause chaotic behavior. The initial state was set to equilibrium and a small perturbation was given to a randomly selected state. A small amount of noise was added to the resulting $d_y=20$ dimensional feature vector. For the input ranging from $0$ to $8$, $1000$ samples were collected on a linear space from the system. The objective is to identify the latent function generating data and perform predictions at $10^5$locations in this interval, ${[0, 8]}$.

\subsection{Datasets used in the illustrative experiment in Section~\ref{sec:exp}, Figure~\ref{fig:overfit}.}
\subsubsection{\texorpdfstring{$\mathrm{ToyData}$}{Lg}}
\label{app:toy_data}
The synthetic dataset $\mathrm{ToyData}$ for the regression task in Figure~\ref{fig:overfit}-(a) is generated using the following nonlinear functions
\begin{align*}
&f_1(x) = \exp{\{\sin (\cos(x)) \sin(\log(1+|x^2-3x|))\}}, \\
& f_2(x) = \log(|\tan(-2x)\cos(2x) +1|)\sin(x).
\end{align*}
We generated $32$ noisy samples for input values in the range of $x\in[0, 12]$. The objective is to estimate the latent functions and perform predictions at $10^5$ locations in this interval, ${[0, 12]}$.
\subsubsection{\texorpdfstring{$\mathrm{vicon2}$}{Lg}}
\label{app:vicon2}
The dataset $\mathrm{vicon2}$ is a subset of the $\mathrm{vicon}$ dataset (\ref{app:vicon}) which includes $6000$ samples from which $5000$ randomly selected samples are used in the test set and $1000$ samples in the training set. $\mathrm{vicon2}$ is used in our numerical simulation presented in Figure~\ref{fig:overfit}-(b).

\subsection{Methods}
\subsubsection{\texorpdfstring{$\mathrm{SGPMC}$}{Lg}}
\label{app:sgpmc}
MCMC for Variational Sparse Gaussian Processes ($\mathrm{SGPMC}$) model of \cite{Hensman2015} using GPflow implementation\footnote{https://github.com/GPflow} with RBF-ARD kernels, Gaussian likelihood, and $1000$ pseudo inputs.
\subsubsection{\texorpdfstring{$\mathrm{SVGP}$}{Lg}}
\label{app:svgp}
The scalable variational Gaussian process ($\mathrm{SVGP}$) model of \cite{Hensman15b} using a GPflow implementation with RBF-ARD kernels, Gaussian likelihood, and $1000$ pseudo inputs.
\subsubsection{\texorpdfstring{$\mathrm{SVIGP}$}{Lg}}
\label{app:svigp}
Stochastic variational GP ($\mathrm{SVIGP}$) model of \cite{hensman2013} using a  GPy\footnote{https://github.com/SheffieldML/GPy} implementation with RBF-ARD kernels, Gaussian likelihood, and $1000$ pseudo inputs.

\end{document}